\documentclass[runningheads]{llncs}

\usepackage{eccv}

\usepackage{eccvabbrv}

\usepackage{graphicx}
\usepackage{booktabs}

\usepackage[accsupp]{axessibility}  

\usepackage{graphicx}
\usepackage{algorithm}
\usepackage{algpseudocode}
\usepackage{multirow}
\usepackage{makecell}
\usepackage{slashbox}
\usepackage{wrapfig}
\usepackage{graphicx}

\usepackage{hyperref}

\usepackage{orcidlink}
\makeatletter
\newcommand{\printfnsymbol}[1]{%
  \textsuperscript{\@fnsymbol{#1}}%
}
\makeatother

\begin{document}

\title{Defect Spectrum: A Granular Look of Large-Scale Defect Datasets with Rich Semantics} 

\titlerunning{Defect Spectrum}


\author{Shuai Yang\inst{1,2}\thanks{These authors contributed equally to this work.} \and
Zhifei Chen\inst{1}\printfnsymbol{1}\and
Pengguang Chen\inst{3} \and
Xi Fang\inst{3} \and
Yixun Liang\inst{1} \and
Shu Liu\inst{3} \and
Yingcong Chen\inst{1,2,4}
}

\authorrunning{S.~Yang, Z.~Chen et al.}

\institute{Hong Kong University of Science and Technology, Guangzhou \and
HKUST(GZ) - SmartMore Joint Lab\\
 \and
SmartMore. Corp\\
\and
Hong Kong University of Science and Technology
}

\maketitle

\begin{abstract}
Defect inspection is paramount within the closed-loop manufacturing system. However, existing datasets for defect inspection often lack the precision and semantic granularity required for practical applications. In this paper, we introduce the Defect Spectrum, a comprehensive benchmark that offers \textbf{precise}, \textbf{semantic-abundant}, and \textbf{large-scale} annotations for a wide range of industrial defects. Building on four key industrial benchmarks, our dataset refines existing annotations and introduces rich semantic details, distinguishing multiple defect types within a single image. With our dataset, we were able to achieve an increase of \textbf{10.74\%} in the Recall rate, and a decrease of \textbf{33.10\%} in the False Positive Rate (FPR) from the industrial simulation experiment. 
Furthermore, we introduce Defect-Gen, a two-stage diffusion-based generator designed to create high-quality and diverse defective images, even when working with limited defective data. The synthetic images generated by Defect-Gen significantly enhance the performance of defect segmentation models, achieving an improvement in mIoU scores up to \textbf{9.85} on Defect-Spectrum subsets. 
Overall, The Defect Spectrum dataset demonstrates its potential in defect inspection research, offering a solid platform for testing and refining advanced models. 
Our project page is in \url{https://envision-research.github.io/Defect_Spectrum/}.

\end{abstract}

\begin{figure}[h]
\begin{center}
\includegraphics[width=0.85\linewidth]{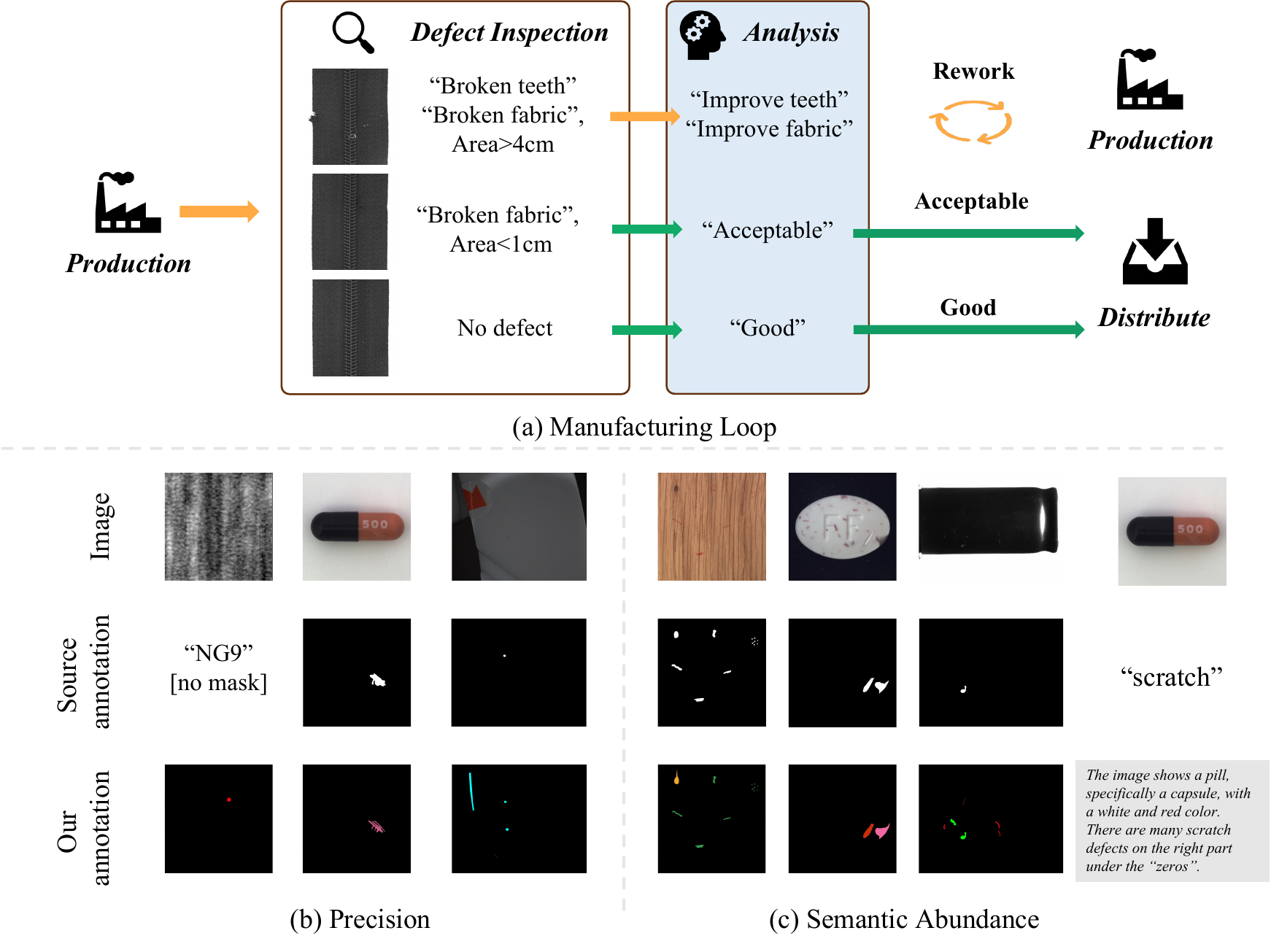}
\end{center}
   \caption{(a) Identifying the size, position, and type of defects is essential for quality control, as it guides the post-processing of products. Major issues, such as misaligned zipper teeth, necessitate factory rework, whereas minor problems, like fabric snags, can lead to different distribution strategies. This approach ensures the maintenance of product quality and enhances the distribution process. (b) Shows our annotation is finer, and includes those that are omitted in the source annotation. (c) Source annotation\cite{bergmann2019mvtec, bai2023vision, wieler2007weakly} ignores multiple defective classes within a single image, while ours provides annotation for each distinct class, shown in different colors. \textbf{Best viewed in color.}} 
\label{fig:teaser}
\end{figure}

\section{Introduction}
\label{introduction}

Industrial manufacturing is a cornerstone of modern society. In an environment where minute imperfections can result in significant failures, ensuring top-tier quality is imperative. Manufacturing predominantly relies on a closed-loop system, encompassing production, defect inspection, filtering, and analysis, as illustrated in Figure~\ref{fig:teaser}.

Within this system, defect inspection plays a pivotal role, interfacing with most stages and ultimately determining product quality. Striking the right balance between identifying defective items and acknowledging sub-optimal ones, based on defect size, position, and type, becomes critical~\cite{DBLP:journals/corr/000116g}. 
Taking the "zipper" defect as an illustrative case. A garment zipper where the teeth are misaligned, as depicted in Figure~\ref{fig:teaser}~(a). This type of defect, although it might seem minor in terms of size or visibility, critically impacts the garment's functionality, necessitating its return to the factory for correction. However, defects located on the fabric, such as minor snags or slight color variations, require careful consideration of their size and impact. Small-scale fabric defects could be classified within an acceptable range, allowing for differentiated distribution strategies that might include selling these products at a discount, thereby maintaining product flow without compromising overall quality standards. Additionally, documenting the category and location of defects can pave the way for predictive maintenance and provide valuable insights for refining product repair processes~\cite{ni2022defect}. 

However, current datasets struggle to meet the intricate practical needs of industrial defect inspection. One notable limitation is the insufficient granularity concerning defect types and locations. For instance, anomaly detection datasets like MVTEC~\cite{bergmann2019mvtec} and AeBAD~\cite{AeBADindustrial} give pixel-level annotations but are restricted to binary masks. Meanwhile, datasets like VISION~\cite{bai2023vision}, though more detailed, occasionally miss or misclassify defect instances.

To address these gaps, we introduce the Defect Spectrum, aiming to offer semantics-abundant, precise, and large-scale annotations for a broad spectrum of industrial defects. This empowers practical defect inspection systems to furnish a more thorough and precise analysis, bolstering automated workflows.
Building on four key industrial benchmarks, Defect Spectrum offers enhanced annotations through a rigorous labeling endeavor.  We have re-evaluated and refined existing defect annotations to ensure a holistic representation. For example, contours of subtle defects, like scratches and pits, are carefully refined for better precision, and missing defects are carefully filled with the help of specialists. Beyond that, our dataset stands out by providing annotations with rich semantics details, distinguishing multiple defect types even within a single image. 
Lastly, we have incorporated descriptive captions for each sample, aiming to integrate the use of Vision Language Models (VLMs) in upcoming studies. During this endeavor, we employ our innovative annotation tool, Defect-Click. It has largely accelerated our labeling process, emphasizing its utility and efficiency, ensuring meticulous labeling even with the extensive scope of our dataset.

Another palpable challenge is the limited number of defective samples in datasets. For instance, in DAGM, there are only $900$ defective images. In MVTEC, although it has $5354$ total images, the defectives among them are merely $1258$. And even the extensive VISION dataset falls short in comparison to natural image datasets like ImageNet~\cite{DenDon09Imagenet} (1 million images) and ADE20k~\cite{zhou2017scene, zhou2019semantic} (20k images).
To address this, we harness the power of generative models, proposing the ``Defect-Gen'', a two-stage diffusion-based generator. Our generator exhibits promising performance in image diversity and quality even with a limited number of training data. We show that these generated data could largely boost the performance of existing models in our Data Spectrum benchmark.  

To summarize, our contributions are listed as follows.
\begin{itemize}
    \item We introduce the Defect Spectrum dataset, designed to enhance defect inspection with its \textbf{semantics-abundant}, \textbf{precise}, and \textbf{large-scale annotations}. Unlike existing datasets, we not only refine existing annotations for a more holistic representation but also introduce rich semantics details. This dataset, building on four key industrial benchmarks, goes beyond binary masks to provide more detailed and precise annotations.
    
    \item We propose the Defect-Gen, a \textbf{two-stage diffusion-based} generator, to tackle the challenges associated with the limited availability of defective samples in datasets. This generator is shown to boost the performance of existing models, by enhancing image diversity and quality even with a limited training set.

    \item We conducted a comprehensive evaluation on our Defect Spectrum dataset, highlighting its versatility and application across various defect inspection challenges. By doing so, we provide a foundation for researchers to evaluate and develop state-of-the-art models tailored for the intricate needs of industrial defect inspection.

\end{itemize}
\section{Related Work}
\label{related_work}

\subsubsection{Industrial Datasets}
There are several well-used datasets for Industrial Defect Inspection: DAGM2007\cite{wieler2007weakly}, AITEX\cite{aitexdataset}, AeBAD\cite{AeBADindustrial}, BeanTech\cite{btad_dataset}, Cotton-SFDG\cite{cottonfabric} and KoektorSDD~\cite{tabernik2020segmentation} offer commonly seen images that cover a wide array of manufacturing materials; MVTEC~\cite{bergmann2019mvtec, bergmann2021mvtec} is a dataset for benchmarking anomaly detection methods with a focus on industrial inspection; VISION V1\cite{bai2023vision} includes a collection of 14 industrial inspection datasets containing multiple objects. A notable shortcoming in the aforementioned industrial datasets is they often lack specificity regarding the defect's type or its precise location. Aiming to refine these issues, we introduce the Defect Spectrum datasets. Further details will be explained in Section~\ref{sec:dataset}.

\subsubsection{Defect-mask Generation}
Defect inspection plays a vital role in various industries, including manufacturing, healthcare, and transportation. Previous attempts based on the traditional computer vision method~\cite{Song2015} have proven to be robust for detecting small defects, but they all suffer from detecting defects in
textures-rich patterns. In recent years, Convolutional Neural Networks(CNNs)~\cite{faghih2016deep, mundt2019meta, guo2021semi} based models are commonly used for defect inspection, but limited availability of real-world defect samples remains a challenge. To mitigate such data-insufficiency issue, traditional methods for synthesizing defect images manually destroy normal samples~\cite{mery2002automated} or adopt Computer-Aided Drawing (CAD)~\cite{mery2005simulation, huang2009template}. 
Deep learning-based approaches are generally effective, but they require large amounts of data. 
GAN-based methods~\cite{niu2020defect, wei2022mask, du2022new, zhang2021defect} are adopted to perform defect sample synthesis for data augmentation. 
DefectGAN adopts an encoder-decoder structure to synthesize defects by mimicking defacement and restoration processes. 
However, it is important to note that GAN-based methods typically require a substantial quantity of real defect data in order to achieve effective results. 
Recent advancements in Diffusion models~\cite{ho2020denoising, dhariwal2021diffusion, nichol2021improved} demonstrated a superior performance in image generation. However, they tend to reproduce existing samples when trained with scarce data, leading to a lack of diversity. Stable Diffusion~\cite{rombach2021highresolution} is one of the most prevailing methods in this field. Nonetheless, it is not applicable to use a pre-trained stable diffusion model when generating masks.
Our proposed approach, on the other hand, is capable of generating defective image-mask pairs with both diversity and high quality, even when trained on limited datasets.

\section{Dataset}
\label{sec:dataset}

\subsection{Datasets Analysis}
In Table~\ref{tab:dataset}, we present an analysis of the Defect Spectrum datasets in comparison with other prevalent industrial datasets. Notably, the DAGM2007 and Cotton-Fabric datasets originally lacked pixel-wise labels, making them less suitable for detailed defect inspection. While datasets like AITEX, AeBAD, BeanTech, and KoektorSDD offer defect masks, they only focus on a limited range of products, offering a restricted number of annotated images and defect categories.

While some high-quality datasets offer a significant volume of images with pixel-level annotations, they are not without their limitations. For instance, there are cases where MVTEC and VISION annotations either miss defects or provide imprecise, coarse labels, as illustrated in Figure~\ref{fig:teaser}(b). Additionally, these datasets commonly merge various defect classes into a single homogeneous category. This shortcoming is particularly apparent in the ``pill'' and ``capacitor'' examples in Figure~\ref{fig:teaser}(c), where the original annotations provide only binary masks that do not differentiate between defects such as ``scratch'', ``crack'', and ``color point''. This approach fails to reflect real-world scenarios, where industrial images frequently exhibit multiple types of defects simultaneously.


To enhance the capabilities for defect detection, Defect Spectrum datasets introduce a comprehensive collection of 3518 high-quality, high-resolution images derived from the aforementioned datasets. These selected images feature a wide variety of objects and defects, ensuring extensive variance and coverage for improved analysis. This curated dataset offers detailed, precise, and diverse category annotations for each image and enriches the data with comprehensive captions to facilitate better contextual understanding. For every product type featured, the Defect Spectrum datasets extend their utility by incorporating realistic synthetic data and their accurate masks, ensuring a thorough and versatile testing ground.

\begin{table}[h]
\centering
\vspace{-0.2in}
\caption{Comparison with real-world manufacturing datasets. Defect Spectrum datasets are the second largest one even though excluding our synthetic data. Defect Spectrum is also the most diverse, semantics-abundant, and precise manufacturing benchmark datasets to date. We use * to represent the amount of synthetic data.}
\begin{tabular}{cccccc}
\hline
\multicolumn{1}{c|}{}                & \begin{tabular}[c]{@{}c@{}}Annotated\\ Defective\\ Images\end{tabular} & \begin{tabular}[c]{@{}c@{}}Defect\\ Type\end{tabular} & \begin{tabular}[c]{@{}c@{}}Pixel-wise\\ Label\end{tabular} & \begin{tabular}[c]{@{}c@{}}Multiple\\ Defective\\ Label\end{tabular} & \begin{tabular}[c]{@{}c@{}}Detailed \\ Caption\end{tabular} \\ \hline
\multicolumn{1}{c|}{AITEX\cite{aitexdataset}}           &  105    &        12         &    \checkmark &                                      &                         \\
\multicolumn{1}{c|}{AeBAD\cite{AeBADindustrial}}           &  346   &       4  &    \checkmark &                                      &                         \\
\multicolumn{1}{c|}{BeanTech\cite{btad_dataset}}        &   290     &    3            &    \checkmark             &                   &                           \\
\multicolumn{1}{c|}{Cotton-Fabric\cite{cottonfabric}}        &   89     &    1            &                &                   &                           \\
\multicolumn{1}{c|}{DAGM2007\cite{wieler2007weakly}}        &   900    &   6  &                  &                                   &                                 \\
\multicolumn{1}{c|}{KolektorSDD2\cite{tabernik2020segmentation}}    &  356           &         1       &   \checkmark       &                            &                          \\
\multicolumn{1}{c|}{MVTec\cite{bergmann2019mvtec}}           &    1258     &        69               &        \checkmark              &                            &                         \\
\multicolumn{1}{c|}{VISION V1\cite{bai2023vision}}       &   4165                   &       44   &   \checkmark &       \checkmark            &                   \\ 
\multicolumn{1}{c|}{VisA \cite{zou2022spotthedifference}}     &   1200                   &       75   &   \checkmark &                   &                   \\ \hline
\multicolumn{1}{c|}{Defect Spectrum} &   \textbf{3518+1920*}      &     \textbf{125}                  &      \checkmark        &     \checkmark                &      \checkmark        \\ \hline
\end{tabular}
\vspace{-0.1in}
\label{tab:dataset}
\end{table}

\subsection{Annotation Improvements} 

Our improvements in annotations are mainly in three aspects: precision, semantics-abundance, and detailed caption.

\subsubsection{Precision}
For datasets that were not annotated or merely had image-wise annotations, we have elevated them to meet our standards. We have enriched these datasets with meticulous pixel-level annotations, delineating defect boundaries and assigning a distinct class label to each type of defect. For those datasets that already possessed pixel-wise masks, we enhanced their precision and rectified any imperfections. We undertook efforts to account for any overlooked defects, ensuring exhaustive coverage. For nuanced defects, such as scratches and pits, we refined the contours to achieve heightened accuracy. 

\subsubsection{Semantics Abundance}
In contrast to datasets that only offer binary defective masks, Defect Spectrum furnishes annotations with more semantic details, identifying multiple defect types within a single image. We identify that there are 552 multiple defective images and provide their multi-class labels.
Moreover, we have re-assessed and fine-tuned the existing defect classes, guaranteeing a more granular and precise categorization. In total, we offer 125 distinct defect classes.

\subsubsection{Detailed Caption}
With the evolution of Vision Language Models (VLMs), we have equipped our datasets by integrating exhaustive captions. It's worth noting that current captioning models, such as BLIP2~\cite{li2023blip2} and LLaVa~\cite{liu2023llava}, often overlook defect information. 
As a remedy, we manually refined the captions from VLMs and furnished detailed descriptions. These narratives not only identify the objects but also elucidate their specific defects. We anticipate that this enhancement will inspire researchers to increasingly leverage VLMs for defect inspection in forthcoming studies.

\subsection{Defect Generation}
To tackle the issue of defects scarcity, we turn to the burgeoning field of generative models. By using the limited available data, we propose a two-staged diffusion-based generator, called the ``Defect-Gen''.

\label{sec:DiffGen_limit}
\begin{wrapfigure}{r}{0.35\textwidth}
    \begin{center}
    \vspace{-0.55in}
    \includegraphics[width=1.\linewidth]{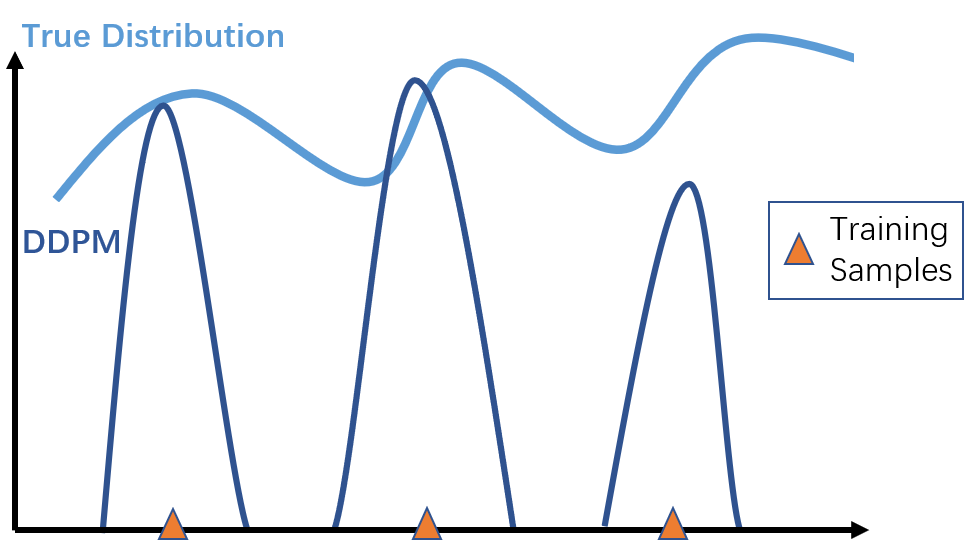}
    \end{center}
    \vspace{-0.25in}
    \caption{DDPM predicts high density around training samples and fails to capture the true data distribution.}
    \vspace{-0.3in}
    \label{distribution}
\end{wrapfigure}
\vspace{-0.17in}
\subsubsection{Background}
Given a set of defective image-mask pairs, we aim to learn a generative model that captures the true data distribution, so that it can generate more pairs to augment the training set. 
We denote the dataset as $\mathcal{D}=\{(I_1, M_1), (I_2, M_2), \dots, (I_N, M_N) \}$, where the image $I_i \in \mathbb{R}^{h\times w \times 3}$ and the mask $M_i \in \{0,n\}^{h\times w \times n}$ refer to the defect image and its corresponding defect mask respectively. $N$ is the number of samples in the training set, which is small in practice. $n$ denotes the number of defective types in the mask images. Specifically, we convert the mask into a one-hot encoding scheme for each channel separately.
We show that with a very small modification, it can generate images with corresponding labels. We perform a channel-wise concatenation between $I$ with the $M$, i.e., $x=I \oplus M$, where $\oplus$ means concatenation and $x\in \mathbb{R}^{h\times w \times n_{total}}$, and $n_{total}=n_{defect}+3$.  We then treat the $x$ as the \textit{input} to train the generator. This improves the usability of the generative model with negligible computational overhead. 
In the following, we term $x$ as ``image'' instead of ``image with label'' for convenience. 

\begin{figure}[h]
\centering
\vspace{-0.2in}
\includegraphics[width=0.95\linewidth]{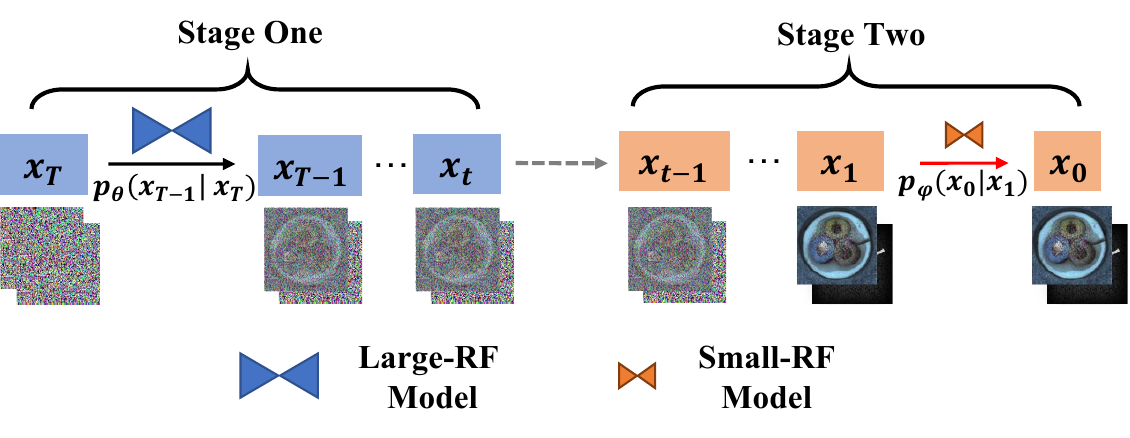}
\vspace{-0.1in}
\caption{The inference process of the two staged diffusion models. The input to the large model $p_{\theta}$ is gaussian noise, after the optimal step is reached, the intermediate results containing global information will be used as the input to the small model $p_{\phi}$.}
\vspace{-0.2in}
\label{pipeline}
\end{figure}

\vspace{-0.1in}
\subsubsection{Few-shot Challenges} 
Note that defect images are difficult to collect in practice, and thus, models have to be trained with very few samples. Under this situation, we observe that the generated results lack diversity. To be specific, models tend to memorize the training set. The reason could be that the generative models such as Diffusion models tend to predict high density around training samples and fail to capture the true data distribution, as depicted in Figure~\ref{distribution}.
\vspace{-0.05in}

\begin{wrapfigure}{r}{0.5\textwidth}
\vspace{-0.5in}
\centering
\includegraphics[width=0.9\linewidth]{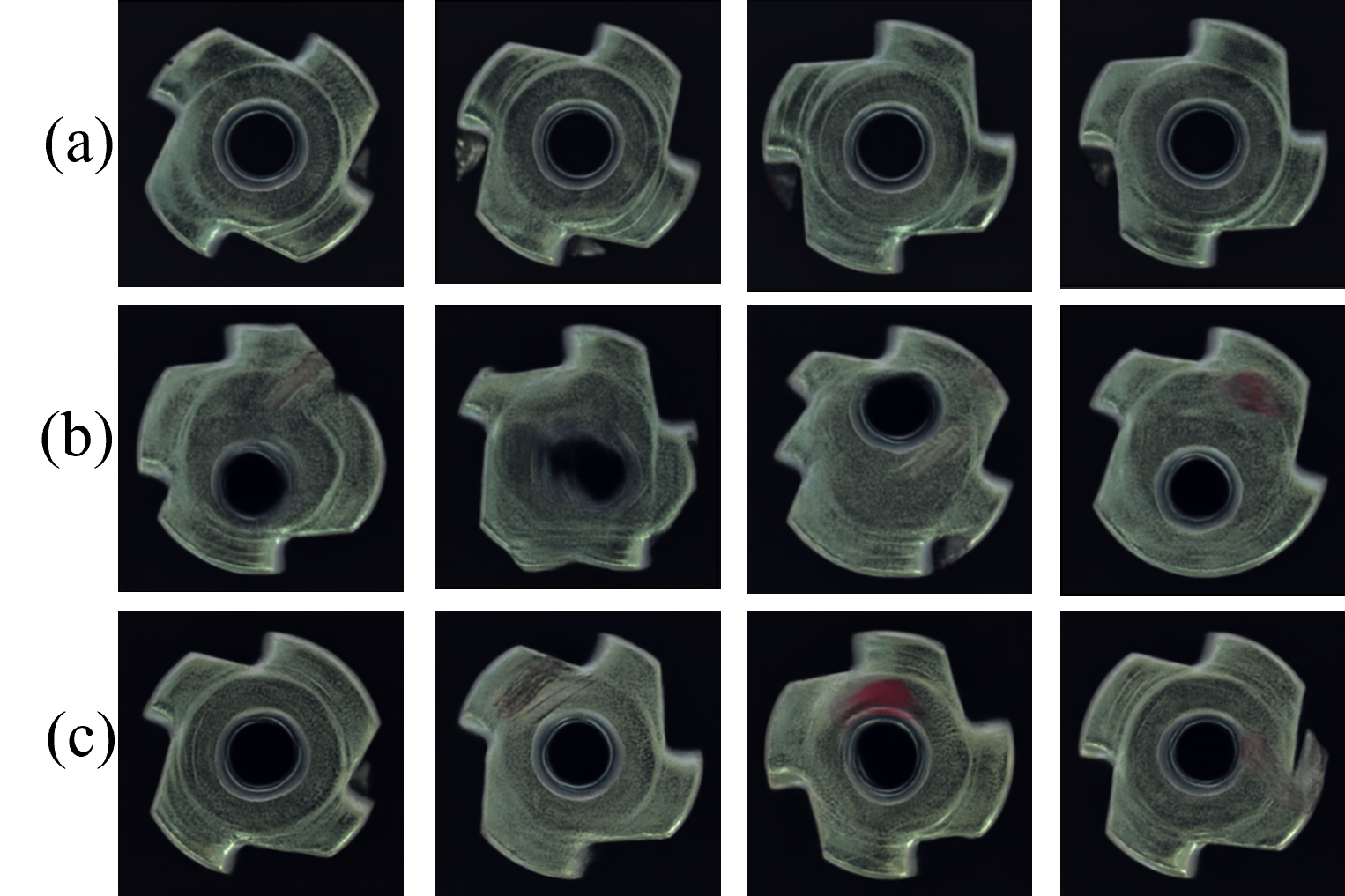}
\vspace{-0.1in}
\caption{The visual cases in (a) demonstrate a lack of diversity in using DDPMs. Cases in (b) demonstrated excessive diversity. (c) shows the generated samples using our framework. we maintained the global structure while introducing local variance.}
\vspace{-0.5in}
\label{metalnut}
\end{wrapfigure}
\vspace{-0.12in}
\subsubsection{Overfitting Issue}
\label{sec:DiffGen_method}
The limitation discussed above is not surprising. In statistical learning theory, it is well-known that the generalization capacity of a classification model is positively related to the sample size and negatively related to the dimension. We can reasonably hypothesize that a similar trend also holds in the diffusion model according to the Vapnik–Chervonenkis theory~\cite{vapnik2015uniform}. In this sense, as the data dimension ($h\times w \times n_{total}$) is much larger than the sample size ($N=25$ in our setting), the vanilla diffusion model suffers severe overfitting. As shown in Figure~\ref{metalnut}~(a), DDPM replicates training cases, leading to low diversity generation.


\subsubsection{Modeling the Patch-level Distribution} 
\begin{wrapfigure}{r}{0.5\textwidth}
\centering
\vspace{-0.1in}
\includegraphics[width=1\linewidth]{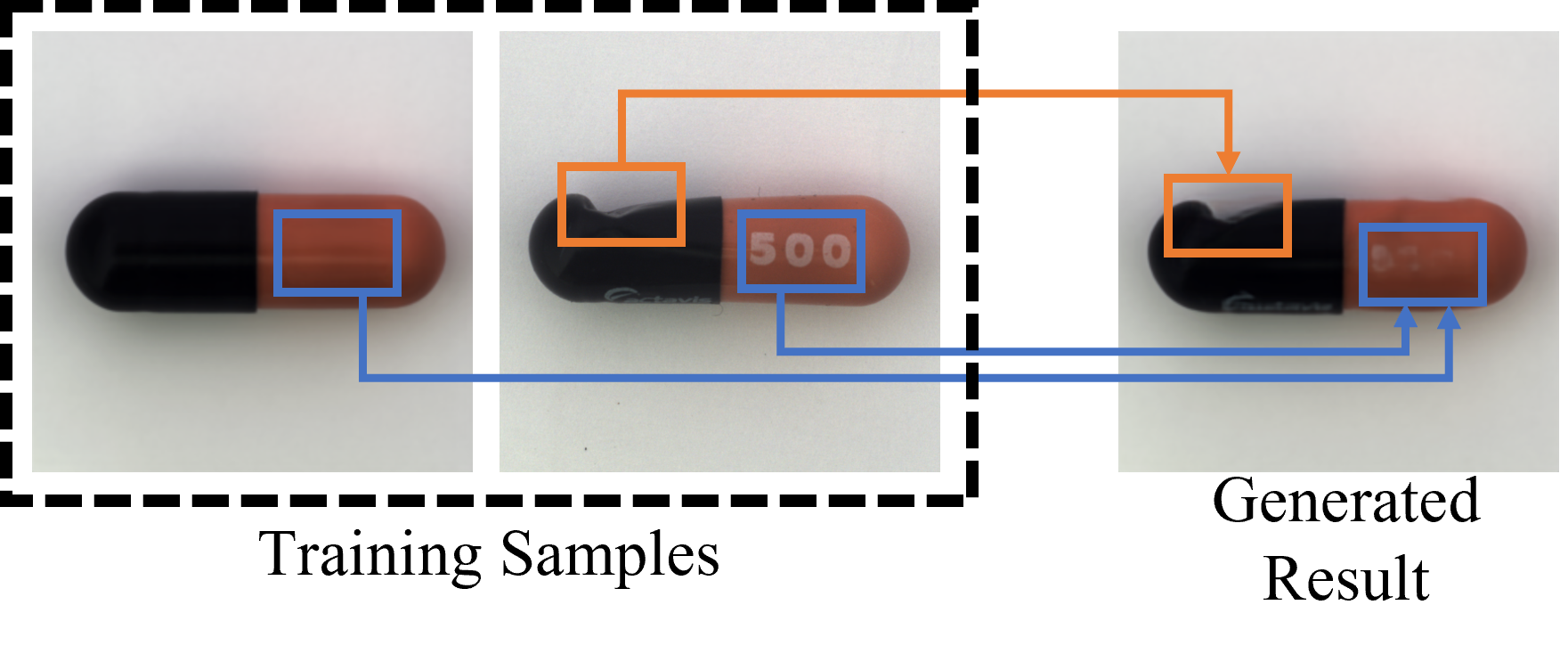}
\vspace{-0.25in}
\caption{The property of patch-level modeling. The right image is generated from the small-receptive-field model, and the two left images are the two most similar images from the training set.}

\label{receptive_field_fusion}
\end{wrapfigure}

To alleviate the aforementioned problem, we propose to model the patch-level distribution instead of the image-level distribution. By treating a patch as one sample, the data dimension ($h_{patch} \times w_{patch} \times n_{total}$) is largely reduced, while the sample size ($N_{patch}$)is significantly increased. This reduces the risk of overfitting. Figure~\ref{receptive_field_fusion}. demonstrates the effectiveness of our strategy.

\subsubsection{Restraining the Receptive Field} 
\label{param_select_0}
Although we can naively replace $x$ with cropped image patches to achieve patch-level modeling, it is hard to use learned patches to reconstruct into a whole image during inference. In other words, if explicitly train a patched generator, we would have to introduce a reconstruction term to merge these patches. Alternatively, we leverage the network architecture to restrain the size of the receptive field to achieve this. Standard U-Net is used in the vanilla diffusion model~\cite{ho2020denoising}. It is composed of a series of down-sampling layers. With the reduced number of down-sampling layers, the output receptive fields gradually decrease. This allows the model to only be visible to small patches on the original images. This strategy does not change the position of each patch in one image and thus has the potential to maintain the whole image. Thus, by using a smaller receptive field, patch-level modeling is achieved. 

\subsubsection{Handling the Global Distortion} \label{param_select_1}

While patch-level modeling is effective in overcoming overfitting, it falls short of representing the global structure of the entire image, leading to unrealistic results. This is shown in Figure~\ref{metalnut}(b). 
To address this issue, we propose a two-stage diffusion process as depicted in Figure~\ref{pipeline}. Our approach is inspired by \cite{choi2022perception}, which reveals that different time steps in the diffusion process correspond to distinct levels of information. In the early stages, coarse geometry information is generated, while in later stages, finer information is produced.

Specifically, we train two models: one with a small receptive field, which we introduced previously, and another with a larger receptive field. During inference, we use the large-receptive-field model to capture the geometry structure in the early steps, and then switch to the small-receptive-field model to generate diverse local patches in the remaining steps. The effectiveness of this strategy is demonstrated in Figure~\ref{metalnut}~(c). Our model has two key hyper-parameters: the switch timestep $u$ and the receptive field of the small model. Both of them can control the trade-off between fidelity and diversity. We use FID to measure the generation fidelity. LPIPS was originally used for measuring the similarity between two images, the lower score indicates a higher similarity and vice versa. In this scenario, to achieve a higher generation diversity with fidelity, we want to maintain a higher LPIPS score with a similar FID score. Due to the page limits, the detailed selection of the switch timestep $u$ and the receptive field of the small model can be found in the Sec.\textcolor[rgb]{ 1,  0,  0}{B} of the Appendix.

\subsection{Auxiliary Annotation Tool}

\begin{figure}[h]
\begin{center}
\includegraphics[width=1.0\linewidth]{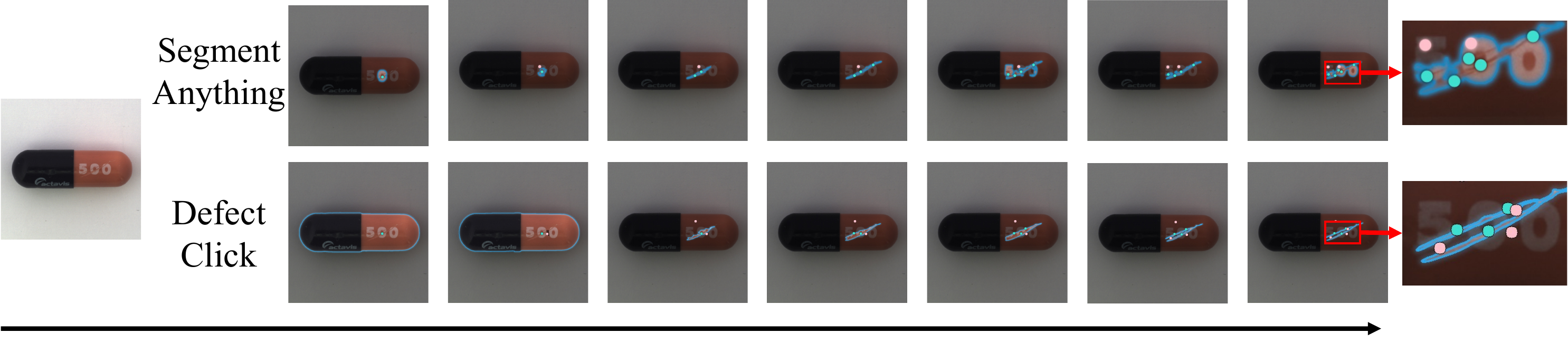}
\end{center}
   \caption{Comparison between Defect Click and Segment Anything~\cite{kirillov2023segany}. Progressively annotating a scratched capsule with human clicks: With our ``Defect-Click'' tool, we can swiftly pinpoint the two scratches. However, when using ``Segment Anything'', it becomes challenging to accurately identify the defects, as shown in the red box. \textbf{Best viewed in color.}}
   \vspace{-0.1in}
   \vspace{-0.05in}
\label{fig:defect-click}
\end{figure}

Annotating pixel masks is an exceptionally demanding task in the labeling domain, especially under the stringent standards of Defect Spectrum. It is not feasible to perform such a task from scratch. To alleviate this challenge, we introduce an auxiliary annotation tool, ``Defect-Click,'' designed to conserve the efforts of our specialists.

Defect-Click is an advanced interactive annotation tool designed to automatically segment defect areas based on a user's click point. Distinct from traditional interactive segmentation methods, Defect-Click utilizes its pretrained knowledge of industrial defects to adeptly pinpoint irregular defective regions. Built upon the Focal-Click framework~\cite{cdnet, focalclick}, we tailored Defect-Click for the industrial defect domain by integrating 21 proprietary labeled datasets, introducing multi-level crop training for small defects, and incorporating edge-sensitive losses during training.  The 21 proprietary labeled datasets consist of defective image-mask pairs for industrial inspection. Multi-level crop training means we rescale the training samples randomly to a resolution of [512, 1024, 1536, 2048, 2560, 3072] and then crop $512\times512$ patches for training. Edge-sensitive losses denote the loss function in Mask2Former~\cite{cheng2022maskedattention}. We use the $loss_{cls} : loss_{mask} : loss_{dice} = 2 : 5 : 5$ in practice. These specialized approaches ensure that Defect-Click significantly outperforms other annotation tools in the industrial dataset domain, as showcased in Figure~\ref{fig:defect-click}. Segment Anything~\cite{kirillov2023segany} struggles to identify the scratch defect, while Defect-Click clearly delineates the defect's contour.

With the assistance of Defect-Click, we can initially obtain a rough defect mask with merely several clicks and subsequently refine its imperfections. On average, this approach has resulted in a time-saving of about 60\%. Even though, this comprehensive annotation project still spans a total of 580 working hours.

\section{Experiments}

\subsection{Benchmarking existing methods}
\label{baseline}
In the realm of industrial defect inspection, there are three primary tasks: defect detection (determining if an image contains a defect), defect classification (identifying the type of defect), and defect segmentation (pinpointing both the boundaries and the type of the defect in the image)~\cite{Carvalho_2022, tang2024incremental, lu2023removing}. Typical defect detection methods such as Patchcore~\cite{roth2022total}, PADIM~\cite{defard2020padim}, and BGAD~\cite{BGAD} emphasize identifying the presence of defects but fall short in discerning defect types. Defect classification methods can determine the type of defect but do not provide information about its location or size. Our Defect Spectrum dataset come with detailed and comprehensive annotations, aiming to solve the most complex task. Consequently, we focus on methods that excel in defect segmentation.

Additionally, due to the confidential nature of many industrial products, transferring data externally is often prohibited. This necessitates models that can operate efficiently on local devices. With this in mind, we have handpicked several SOTA segmentation methods and adapted them to a lightweight version.
Our baseline includes UNet - small~\cite{unet2015}, ResNet18~\cite{he2016deep} - PSPNet~\cite{zhao2017pyramid}, ResNet18 - DeepLabV3+~\cite{chen2018encoder}, HRNetV2W18 - small~\cite{wang2020deep}, BiseNetV2~\cite{yu2021bisenet}, ViT - Tiny~\cite{dosovitskiy2021image}- Segmenter~\cite{strudel2021segmenter}, Segformer - MiT - B0~\cite{xie2021segformer}, and HRNet - Mask2Former~\cite{cheng2022maskedattention}. The models are abbreviated as follows: UNet (UNet - small), PSP (ResNet18 - PSPNet), DL (ResNet18 - DeepLabV3+), HR (HRNetw18small), Bise (BiseNetV2), V-T (ViT-Tiny - Segmenter), M-B0 (Segformer - MiT-B0), and M2F (HRNet - Mask2Former). 


We present a comprehensive evaluation of the above methods on each sub-set of our Defect Spectrum benchmark. For the performance metric, we choose the mean Intersection over Union (mIoU). Results are shown in Table~\ref{tab:downstream_eval}. 
The consistent performance of DeepLabV3+ across multiple datasets suggests that it's a robust model for various types of defect segmentation tasks. 
The Transformer-based models seem to be particularly effective for Cotton-Fabric. This might be due to the inherent advantages of Transformers in capturing long-range semantic information, which could be commonly found in ``Cotton-Fabric''.
Performance varies across models for different categories, suggesting no universal solution. Model selection should consider dataset specifics. Some datasets challenge all models, highlighting a need for more research.

\begin{table}[h]
\centering
\caption{Quantitative comparison of various defect segmentation methods across different Defect Spectrum reannotated datasets. Results reflect the mIoU. We highlight the best mIoU of each dataset with red color. ``DS'' is abbreviated for Defect Spectrum.}
\vspace{0.1in}
\begin{tabular}{c|c|ccccc|ccc}
\hline
& & \multicolumn{5}{c|}{CNNs} & \multicolumn{3}{c}{Transformers} \\
& & UNet & PSP & DL & HR & \multicolumn{1}{c|}{Bise} & V-T & M-B0 & M2F \\ \hline
\multirow{15}{*}{\begin{tabular}[c]{@{}c@{}}Defect\\ Spectrum\\ (MVTec)\end{tabular}}
& bottle & 43.44 & 50.20 & 56.53 & 45.02 & 44.92 & \textcolor[rgb]{ 1,  0,  0}{69.71} & 40.88 & 53.20 \\
& cable & 47.95 & 52.50 & 52.59 & 50.39 & 45.24 & 54.51 & \textcolor[rgb]{ 1,  0,  0}{58.31} & 49.72 \\
& capsule & 28.05 & 29.59 & 35.49 & 34.02 & 28.30 & 33.94 & \textcolor[rgb]{ 1,  0,  0}{38.95} & 26.91 \\
& carpet & 50.91 & \textcolor[rgb]{ 1,  0,  0}{53.76} & 53.75 & 47.28 & 44.52 & 43.70 & 38.45 & 47.34 \\
& grid & 37.06 & \textcolor[rgb]{ 1,  0,  0}{42.86} & 41.18 & 30.97 & 33.89 & 40.08 & 18.86 & 24.81 \\
& h\_nut & 58.84 & 56.87 & \textcolor[rgb]{ 1,  0,  0}{61.78} & 59.31 & 57.53 & 55.07 & 59.60 & 56.72 \\
& leather & 57.56 & \textcolor[rgb]{ 1,  0,  0}{61.42} & 54.56 & 55.45 & 57.89 & 47.85 & 50.80 & 53.96 \\
& m\_nut & 49.18 & 46.99 & 51.08 & 48.76 & \textcolor[rgb]{ 1,  0,  0}{55.51} & 54.68 & 48.89 & 39.43 \\
& pill & 35.81 & 36.38 & 33.83 & 29.30 & 27.23 & 42.65 & \textcolor[rgb]{ 1,  0,  0}{46.35} & 27.14 \\
& screw & 31.87 & \textcolor[rgb]{ 1,  0,  0}{38.77} & 33.36 & 29.66 & 19.01 & 22.54 & 19.26 & 21.89 \\
& tile & 85.49 & 82.51 & 83.02 & \textcolor[rgb]{ 1,  0,  0}{85.66} & 84.21 & 78.29 & 79.14 & 83.04 \\
& t\_brush & 23.96 & 25.25 & 25.16 & 26.25 & 25.58 & \textcolor[rgb]{ 1,  0,  0}{33.30} & 32.22 & 28.26 \\
& tran. & 40.37 & 44.02 & \textcolor[rgb]{ 1,  0,  0}{58.23} & 44.50 & 45.97 & 53.60 & 41.13 & 50.87 \\
& wood & 72.69 & 67.93 & 68.21 & 69.00 & 67.81 & 62.62 & \textcolor[rgb]{ 1,  0,  0}{73.02} & 59.66 \\
& zipper & 54.83 & \textcolor[rgb]{ 1,  0,  0}{60.95} & 58.03 & 55.87 & 47.15 & 51.69 & 60.12 & 49.47 \\
& \textbf{mean} & \textbf{49.88}& \textbf{51.41}  &  \textbf{\textcolor[rgb]{ 1,  0,  0}{51.58}}  & \textbf{47.99} & \textbf{45.40} & \textbf{49.31} & \textbf{46.45}  & \textbf{45.70} \\ \hline
\multirow{5}{*}{\begin{tabular}[c]{@{}c@{}}Defect\\ Spectrum\\ (VISION)\end{tabular}}
& Capa. & 57.04 & 54.01 & 52.75 & 54.56 & 54.36 & 56.30 & \textcolor[rgb]{ 1,  0,  0}{59.29} & 57.27 \\
& Console & \textcolor[rgb]{ 1,  0,  0}{35.32} & 30.77 & 32.67 & 31.70 & 30.45 & 22.48 & 25.64 & 32.50 \\
& Ring & 52.37 & 56.16 & 56.97 & 60.17 & 54.06 & 44.27 & 52.21 & \textcolor[rgb]{ 1,  0,  0}{62.09} \\
& Screw & 53.13 & 53.91 & \textcolor[rgb]{ 1,  0,  0}{55.20} & 52.46 & 51.76 & 36.87 & 47.54 & 52.05  \\
& Wood & 64.75 & 66.80 & 66.72 & 66.79 & \textcolor[rgb]{ 1, 0, 0}{67.40} & 53.34 & 63.43 & 66.70  \\
& \textbf{mean} & \textbf{52.52} & \textbf{52.33} & \textbf{52.86} & \textbf{53.14} & \textbf{51.61} & \textbf{42.65} & \textbf{49.62} & \textbf{\textcolor[rgb]{ 1,  0,  0}{54.12}} \\ \hline
\multicolumn{2}{c|}{DS-DAGM2007} & 85.89 & 85.14 & \textcolor[rgb]{ 1,  0,  0}{86.82} & 84.02 & 83.14 & 52.42 & 83.06 & 85.56\\ \hline
\multicolumn{2}{c|}{DS-Cotton-Fabric} & 39.03 & 48.73 & 47.55 & 41.13 & 46.82 & 51.29 & 50.52 & \textcolor[rgb]{ 1,  0,  0}{64.09}\\
\hline
\end{tabular}
\label{tab:downstream_eval}
\end{table}

\subsection{Generation Quality} 

\vspace{-0.2in}

\begin{figure}[h]
\centering
\includegraphics[width=1.0\linewidth]{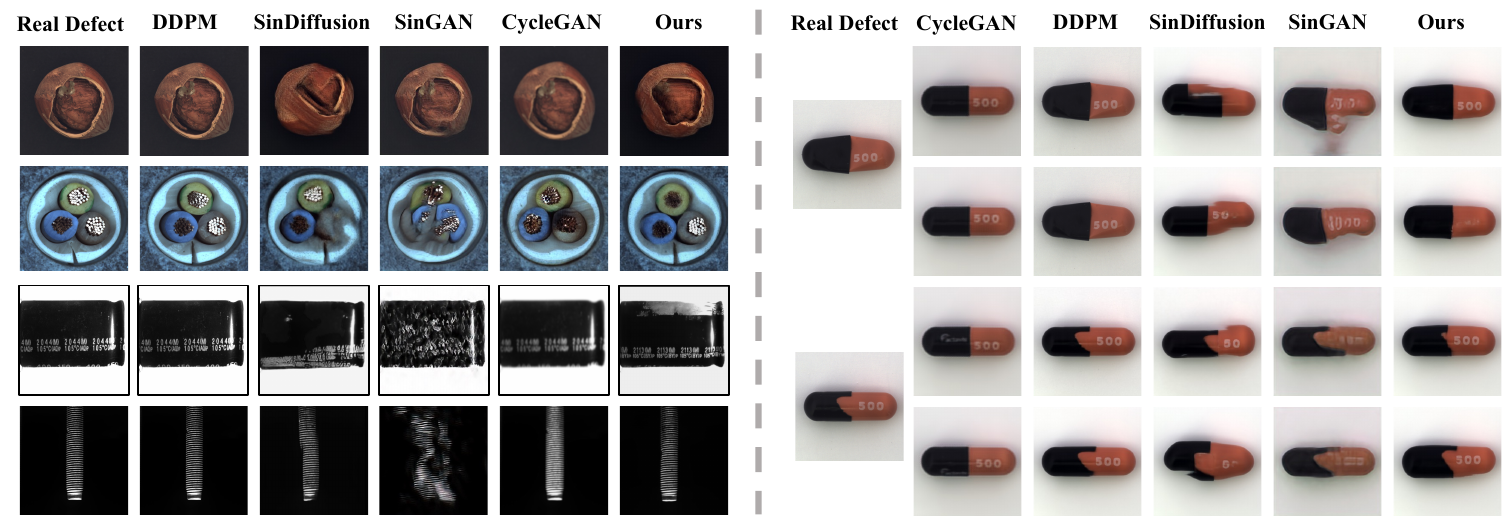}
\caption{Qualitative comparison of our method with other image synthesis methods. On the left-hand side, we compared different objects across different datasets to demonstrate the high fidelity of our generation method. On the right-hand side, we show our method can exhibit diversity while maintaining high quality. \textbf{Best viewed in color.}}
\label{qualitative}
\end{figure}

Figure~\ref{qualitative} provides a qualitative comparison between our generation results and those from other synthesis methods.
On the left-hand side, we present different objects to demonstrate the high fidelity of our method.
On the right-hand side, we used the two images shown in the ``Real defect'' to generate samples to demonstrate our high diversity. 
We observe that the generated models from CycleGAN\cite{CycleGAN2017} and DDPM\cite{ho2020denoising} completely failed to learn a diverse defect pattern and thus failed to generate samples with diversity by producing mere duplicates of the training set. On the other hand, sinDiffusion~\cite{wang2022sindiffusion} and SinGAN~\cite{rottshaham2019singan} can produce diverse samples but are not visually realistic. More visual cases, including other classes, can be found in the supplementary file. 
Figure~\ref{fig:generated_pairs} displays the image-mask pairs we generated. Our images are of high quality, and the corresponding masks align well with them.

\vspace{-0.1in}

\begin{figure}[h]
\centering
\includegraphics[width=0.8\linewidth]{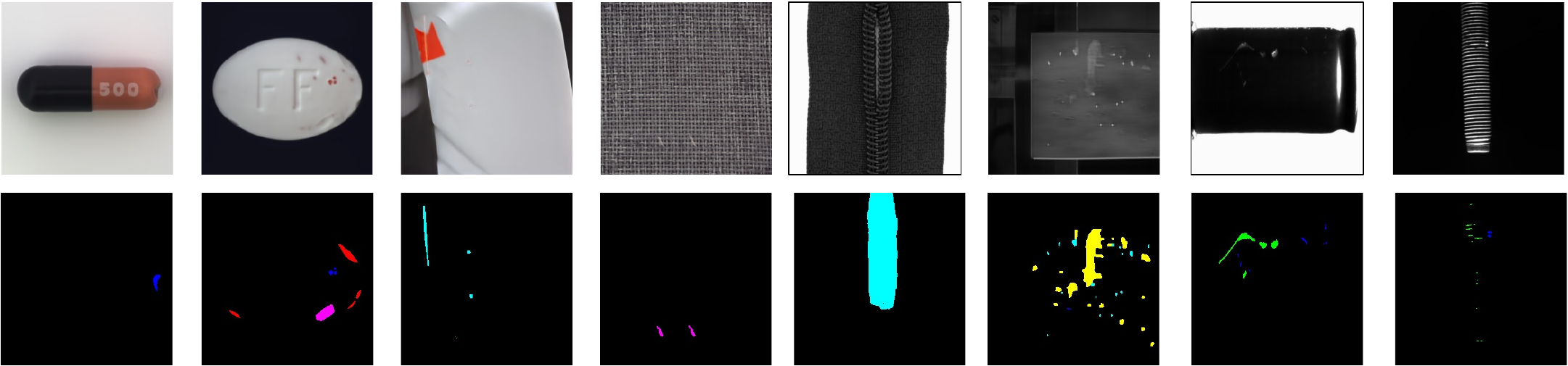}
\caption{Qualitative results of our proposed defect generation method. Generated images demonstrated rich semantics, exhibiting high quality. Generated masks precisely reflected defect areas. \textbf{Best viewed in color.}}
\label{fig:generated_pairs}
\end{figure}

\subsection{Synthetic Data for Performance Boost}

\subsubsection{Boosting SOTA methods with Synthetic Data}
Results in Table~\ref{table:with_syn} show a large performance increase in both DS-MVTec and DS-Cotton datasets, the increase is comparatively smaller in the DS-VISION dataset, however, such increase is demonstrated in each of the sub-classes.
We do not generate extra data for DS-DAGM2007, since it is already a synthetic dataset. The result demonstrates the effectiveness of our synthetic data.
We also compared with other generation methods in the ability to boost performance. Detailed comparisons can be found in the supplementary file. 

\begin{table}[h]
\centering
\caption{Performance (mIoU) comparison between models trained with and without synthetic data. The bolded text indicates results with synthetic data. ``DS'' is abbreviated for Defect Spectrum.}
\begin{tabular}{c|ccc}
\hline
              & {DS-MVTec} & DS-VISION & DS-Cotton \\ \hline
DeepLabV3+ & 51.58/\textbf{55.55}  & 52.33/\textbf{53.46}  & 48.73/\textbf{58.58} \\
Mask2Former  & 45.70/\textbf{50.16}  & 54.12/\textbf{55.47}  & 64.09/\textbf{65.39} \\ 
MiT-B0  & 46.45/\textbf{56.21}  & 49.62/\textbf{50.75}  & 50.52/\textbf{55.86} \\ \hline
\end{tabular}
\label{table:with_syn}
\vspace{-0.1in}
\end{table}


\subsubsection{Impact of Synthetic Data}
\begin{figure}[h]
\vspace{-0.2in}

\centering
\includegraphics[width=0.9\linewidth]{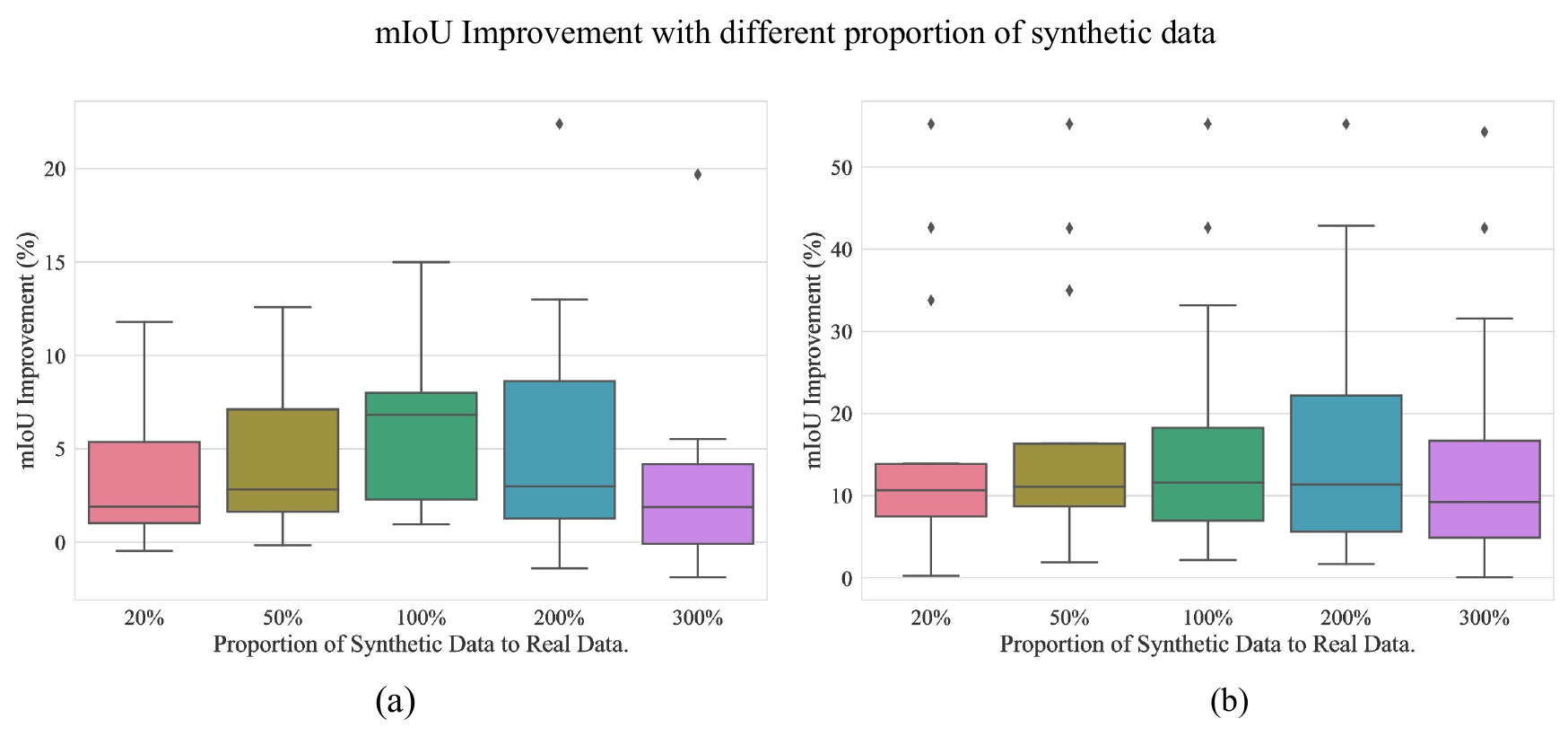}
\caption{Improvement in mIoU with different proportions of synthetic Data. This experiment is done on Defect Spectrum (MVTec) with DeeplabV3+ and MiT-B0 shown as (a) and (b) respectively.}
\label{fig:syn_miou}
\end{figure}

In Figure~\ref{fig:syn_miou}, we delve deeper into the impact of varying the quantity of our synthetic data on model performance. Figure~\ref{fig:syn_miou} (a) shows the performance improvement over different quantities of synthetic data using DeepLabV3+. Interestingly, we found that the transformer-based model (MiT-B0) benefits much more with synthetic data than CNN-based models, as shown in Figure~\ref{fig:syn_miou} (b).

\vspace{-0.2in}

When using synthetic data that is 20\% of the size of the original training set, there is an enhancement in the results. Additionally, it's worth noting that the optimal amount of synthetic data required can vary based on the specific category of images. When using synthetic data that is 200\% of the size of the original training set, there is an enhancement in the performance, but results in greater variance. Additionally, the performance starts to decrease after reaching the 300\%.
On a holistic scale, integrating 100\% of the synthetic data appears to be a reasonable choice.

\subsection{Comparison between original and Defect Spectrum dataset}


\vspace{-0.1in}
\begin{table}[h]
\centering
\caption{The quality control benchmark for the objects to be inspected. We show Zipper, pill, and wood as example classes.}
\begin{tabular}{c|c}
Example classes    & Standard For Benign Products \\ \hline
Zipper &  No defect on teeth; Fabric defect $<$ 4800 pixels        \\
Pill    &  No cracks; Contamination $<$ 4000 pixels; Color stains $<$ 300 pixels          \\
Wood    &  No scratch; No dent; Impurities $<$ 250 pixels; Stain $<$ 1000 pixels   \\
\end{tabular}
\label{table:quality_control}
\end{table}

\begin{wraptable}{r}{0.65\textwidth} 
\vspace{-0.1in}
\centering
\caption{Comparison of original annotation and defect spectrum annotation in the simulation experiment of manufacturing production. Metrics are reported in image level recall rate and false positive rate (FPR).}
\label{table:recall_and_fpr}
\begin{tabular}{lcc} 
\toprule
Method & $\uparrow$ Recall (\%) & $\downarrow$ FPR (\%) \\ 
\midrule
Original & 85.33 & 49.60 \\
Defect Spectrum (DS) & \textbf{96.07} & \textbf{16.50} \\ 
\bottomrule
\end{tabular}
\vspace{-0.3in}
\end{wraptable}

The enhancement of our dataset contains two significant modifications: 1) the expansion to include more defect classes, and 2) the improvement of annotation accuracy in both training and validation sets. Given these significant changes, it becomes impractical to directly assess the performance (like computing mIoU) of our refined model on the original ground truth, and vice versa.

To objectively assess our dataset's superiority, we design a simulation experiment that mirrors real-world manufacturing processes. Manufacturing experts are invited to set a quality control benchmark for the inspected items, as detailed in Table~\ref{table:quality_control}. This benchmark specifies unacceptable critical defects and establishes a threshold for minor defects—for instance, any teeth defect in a zipper is intolerable, whereas only extensive fabric defects in a zipper are considered detrimental.
Following these criteria, we classify the validation samples as either benign or defective.
We then train two segmentation models of the same architecture: one on our refined dataset and the other on the original dataset.
Utilizing these segmentation results and the established criteria, we calculated the image-level recall rate ($\frac{TP}{FN + TP}$) and the false positive rate ($\frac{FP}{TN + FP}$). A superior recall rate denotes more effective defective product identification, whereas a reduced false positive rate indicates fewer benign products mistakenly flagged as defective. As shown in Table~\ref{table:recall_and_fpr}, the model trained with our refined annotations outperforms the one trained with the original dataset in terms of recall rate and false positive rate, thus enhancing product profitability without compromising quality.

Furthermore, we conducted qualitative comparisons across all subsets within the Defect Spectrum. We compared the annotations from the original dataset with those from our refined dataset. Additionally, we evaluated the segmentation model's masks based on the original dataset annotations against those derived from our refined dataset.
Both the masks and the annotations exhibit greater accuracy and improved differentiation among defect types in comparison to the original dataset. Visual examples of these annotations and segmentation masks are included in the appendix for further reference.



\vspace{-0.15in}
\section{Conclusion}
In conclusion, our Defect Spectrum dataset, complemented by the Defect-Gen generator, addresses critical gaps in industrial defect inspection. 
By providing Semantics-abundant, precise, and large-scale annotations, our contributions will foster advancements in defect inspection methodologies. The potential integration of Vision Language Models, the practical value of labeling assistant Defect-Click, coupled with the Defect-Gen's capability to mitigate data scarcity, sets the stage for more robust defect inspection systems in the future. 


%
%
\bibliographystyle{splncs04}
\bibliography{main}

\begin{thebibliography}{10}
\providecommand{\url}[1]{\texttt{#1}}
\providecommand{\urlprefix}{URL }
\providecommand{\doi}[1]{https://doi.org/#1}

\bibitem{bai2023vision}
Bai, H., Mou, S., Likhomanenko, T., Cinbis, R.G., Tuzel, O., Huang, P., Shan,
  J., Shi, J., Cao, M.: Vision datasets: A benchmark for vision-based
  industrial inspection. arXiv preprint arXiv:2306.07890  (2023)

\bibitem{bergmann2021mvtec}
Bergmann, P., Batzner, K., Fauser, M., Sattlegger, D., Steger, C.: The mvtec
  anomaly detection dataset: a comprehensive real-world dataset for
  unsupervised anomaly detection. International Journal of Computer Vision
  \textbf{129}(4),  1038--1059 (2021)

\bibitem{bergmann2019mvtec}
Bergmann, P., Fauser, M., Sattlegger, D., Steger, C.: Mvtec ad--a comprehensive
  real-world dataset for unsupervised anomaly detection. In: Proceedings of the
  IEEE/CVF conference on computer vision and pattern recognition. pp.
  9592--9600 (2019)

\bibitem{Carvalho_2022}
Carvalho, P., Durupt, A., Grandvalet, Y.: A Review of Benchmarks for Visual
  Defect Detection in the Manufacturing Industry, p. 1527–1538. Springer
  International Publishing (Sep 2022). \doi{10.1007/978-3-031-15928-2_133},
  \url{http://dx.doi.org/10.1007/978-3-031-15928-2_133}

\bibitem{chen2018encoder}
Chen, L.C., Zhu, Y., Papandreou, G., Schroff, F., Adam, H.: Encoder-decoder
  with atrous separable convolution for semantic image segmentation. In:
  Proceedings of the European conference on computer vision (ECCV). pp.
  801--818 (2018)

\bibitem{cdnet}
Chen, X., Zhao, Z., Yu, F., Zhang, Y., Duan, M.: Conditional diffusion for
  interactive segmentation. In: ICCV (2021)

\bibitem{focalclick}
Chen, X., Zhao, Z., Zhang, Y., Duan, M., Qi, D., Zhao, H.: Focalclick: Towards
  practical interactive image segmentation  (2022)

\bibitem{cheng2022maskedattention}
Cheng, B., Misra, I., Schwing, A.G., Kirillov, A., Girdhar, R.:
  Masked-attention mask transformer for universal image segmentation (2022)

\bibitem{choi2022perception}
Choi, J., Lee, J., Shin, C., Kim, S., Kim, H., Yoon, S.: Perception prioritized
  training of diffusion models. In: Proceedings of the IEEE/CVF Conference on
  Computer Vision and Pattern Recognition (2022)

\bibitem{defard2020padim}
Defard, T., Setkov, A., Loesch, A., Audigier, R.: Padim: a patch distribution
  modeling framework for anomaly detection and localization (2020)

\bibitem{DenDon09Imagenet}
Deng, J., Dong, W., Socher, R., Li, L.J., Li, K., Fei-Fei, L.: Imagenet: A
  large-scale hierarchical image database. In: Computer Vision and Pattern
  Recognition, 2009. CVPR 2009. IEEE Conference on. pp. 248--255. IEEE (2009),
  \url{https://ieeexplore.ieee.org/abstract/document/5206848/}

\bibitem{dhariwal2021diffusion}
Dhariwal, P., Nichol, A.: Diffusion models beat gans on image synthesis.
  Advances in neural information processing systems  \textbf{34},  8780--8794
  (2021)

\bibitem{dosovitskiy2021image}
Dosovitskiy, A., Beyer, L., Kolesnikov, A., Weissenborn, D., Zhai, X.,
  Unterthiner, T., Dehghani, M., Minderer, M., Heigold, G., Gelly, S.,
  Uszkoreit, J., Houlsby, N.: An image is worth 16x16 words: Transformers for
  image recognition at scale (2021)

\bibitem{du2022new}
Du, Z., Gao, L., Li, X.: A new contrastive gan with data augmentation for
  surface defect recognition under limited data. IEEE Transactions on
  Instrumentation and Measurement  (2022)

\bibitem{faghih2016deep}
Faghih-Roohi, S., Hajizadeh, S., N{\'u}{\~n}ez, A., Babuska, R., De~Schutter,
  B.: Deep convolutional neural networks for detection of rail surface defects.
  In: 2016 International joint conference on neural networks (IJCNN). pp.
  2584--2589 (2016)

\bibitem{guo2021semi}
Guo, J., Wang, Q., Li, Y.: Semi-supervised learning based on convolutional
  neural network and uncertainty filter for fa{\c{c}}ade defects
  classification. Computer-Aided Civil and Infrastructure Engineering pp.
  302--317 (2021)

\bibitem{he2016deep}
He, K., Zhang, X., Ren, S., Sun, J.: Deep residual learning for image
  recognition. In: Proceedings of the IEEE conference on computer vision and
  pattern recognition. pp. 770--778 (2016)

\bibitem{ho2020denoising}
Ho, J., Jain, A., Abbeel, P.: Denoising diffusion probabilistic models.
  Advances in Neural Information Processing Systems  \textbf{33},  6840--6851
  (2020)

\bibitem{huang2009template}
Huang, Q., Wu, Y., Baruch, J., Jiang, P., Peng, Y.: A template model for defect
  simulation for evaluating nondestructive testing in x-radiography. IEEE
  Transactions on Systems, Man, and Cybernetics-Part A: Systems and Humans
  \textbf{39},  466--475 (2009)

\bibitem{cottonfabric}
Incorporated, C.: Standard fabric defect glossary (2023), uRL:
  \url{https://www.cottoninc.com/quality-products/textile-resources/fabric-defect-glossary}

\bibitem{kirillov2023segany}
Kirillov, A., Mintun, E., Ravi, N., Mao, H., Rolland, C., Gustafson, L., Xiao,
  T., Whitehead, S., Berg, A.C., Lo, W.Y., Doll{\'a}r, P., Girshick, R.:
  Segment anything. arXiv:2304.02643  (2023)

\bibitem{li2023blip2}
Li, J., Li, D., Savarese, S., Hoi, S.: {BLIP-2:} bootstrapping language-image
  pre-training with frozen image encoders and large language models. In: ICML
  (2023)

\bibitem{liu2023llava}
Liu, H., Li, C., Wu, Q., Lee, Y.J.: Visual instruction tuning (2023)

\bibitem{lu2023removing}
Lu, F., Yao, X., Fu, C.W., Jia, J.: Removing anomalies as noises for industrial
  defect localization. In: Proceedings of the IEEE/CVF International Conference
  on Computer Vision. pp. 16166--16175 (2023)

\bibitem{mery2005simulation}
Mery, D., Hahn, D., Hitschfeld, N.: Simulation of defects in aluminium castings
  using cad models of flaws and real x-ray images. Insight-Non-Destructive
  Testing and Condition Monitoring pp. 618--624 (2005)

\bibitem{mery2002automated}
Mery, D., Filbert, D.: Automated flaw detection in aluminum castings based on
  the tracking of potential defects in a radioscopic image sequence. IEEE
  Transactions on Robotics and Automation  \textbf{18}(6),  890--901 (2002)

\bibitem{btad_dataset}
Mishra, P., Verk, R., Fornasier, D., Piciarelli, C., Foresti, G.L.: {VT-ADL}: A
  vision transformer network for image anomaly detection and localization. In:
  30th IEEE/IES International Symposium on Industrial Electronics (ISIE) (June
  2021)

\bibitem{mundt2019meta}
Mundt, M., Majumder, S., Murali, S., Panetsos, P., Ramesh, V.: Meta-learning
  convolutional neural architectures for multi-target concrete defect
  classification with the concrete defect bridge image dataset. In: Proceedings
  of the IEEE/CVF Conference on Computer Vision and Pattern Recognition. pp.
  11196--11205 (2019)

\bibitem{ni2022defect}
Ni, C., Yang, K., Xia, X., Lo, D., Chen, X., Yang, X.: Defect identification,
  categorization, and repair: Better together (2022)

\bibitem{nichol2021improved}
Nichol, A.Q., Dhariwal, P.: Improved denoising diffusion probabilistic models.
  In: International Conference on Machine Learning. pp. 8162--8171. PMLR (2021)

\bibitem{niu2020defect}
Niu, S., Li, B., Wang, X., Lin, H.: Defect image sample generation with gan for
  improving defect recognition. IEEE Transactions on Automation Science and
  Engineering  \textbf{17}(3),  1611--1622 (2020)

\bibitem{rombach2021highresolution}
Rombach, R., Blattmann, A., Lorenz, D., Esser, P., Ommer, B.: High-resolution
  image synthesis with latent diffusion models (2021)

\bibitem{unet2015}
Ronneberger, O., Fischer, P., Brox, T.: U-net: Convolutional networks for
  biomedical image segmentation (2015)

\bibitem{roth2022total}
Roth, K., Pemula, L., Zepeda, J., Schölkopf, B., Brox, T., Gehler, P.: Towards
  total recall in industrial anomaly detection (2022)

\bibitem{rottshaham2019singan}
Rott~Shaham, T., Dekel, T., Michaeli, T.: Singan: Learning a generative model
  from a single natural image. In: Computer Vision (ICCV), IEEE International
  Conference on (2019)

\bibitem{aitexdataset}
Silvestre-Blanes, J., Albero-Albero, T., Miralles, I., Pérez-Llorens, R.,
  Moreno, J.: A public fabric database for defect detection methods and
  results. Autex Research Journal  \textbf{19}(4),  363--374 (2019).
  \doi{doi:10.2478/aut-2019-0035}, \url{https://doi.org/10.2478/aut-2019-0035}

\bibitem{Song2015}
Song, W., Chen, T., Gu, Z., Gai, W., Huang, W., Wang, B.: Wood materials
  defects detection using image block percentile color histogram and
  eigenvector texture feature. In: Proceedings of the First International
  Conference on Information Sciences, Machinery, Materials and Energy. Atlantis
  Press (2015). \doi{10.2991/icismme-15.2015.163},
  \url{https://doi.org/10.2991/icismme-15.2015.163}

\bibitem{strudel2021segmenter}
Strudel, R., Garcia, R., Laptev, I., Schmid, C.: Segmenter: Transformer for
  semantic segmentation. In: Proceedings of the IEEE/CVF international
  conference on computer vision. pp. 7262--7272 (2021)

\bibitem{tabernik2020segmentation}
Tabernik, D., {\v{S}}ela, S., Skvar{\v{c}}, J., Sko{\v{c}}aj, D.:
  Segmentation-based deep-learning approach for surface-defect detection.
  Journal of Intelligent Manufacturing  \textbf{31}(3),  759--776 (2020)

\bibitem{tang2024incremental}
Tang, J., Lu, H., Xu, X., Wu, R., Hu, S., Zhang, T., Cheng, T.W., Ge, M., Chen,
  Y.C., Tsung, F.: An incremental unified framework for small defect
  inspection. In: 18th European Conference on Computer Vision (ECCV) (2024),
  \url{https://github.com/jqtangust/IUF}

\bibitem{vapnik2015uniform}
Vapnik, V.N., Chervonenkis, A.Y.: On the uniform convergence of relative
  frequencies of events to their probabilities. Measures of complexity:
  festschrift for alexey chervonenkis  (2015)

\bibitem{DBLP:journals/corr/000116g}
Wagner, S.: A literature survey of the quality economics of defect-detection
  techniques. CoRR  \textbf{abs/1612.04590} (2016),
  \url{http://arxiv.org/abs/1612.04590}

\bibitem{wang2020deep}
Wang, J., Sun, K., Cheng, T., Jiang, B., Deng, C., Zhao, Y., Liu, D., Mu, Y.,
  Tan, M., Wang, X., et~al.: Deep high-resolution representation learning for
  visual recognition. IEEE transactions on pattern analysis and machine
  intelligence  \textbf{43}(10),  3349--3364 (2020)

\bibitem{wang2022sindiffusion}
Wang, W., Bao, J., Zhou, W., Chen, D., Chen, D., Yuan, L., Li, H.:
  Sindiffusion: Learning a diffusion model from a single natural image. arXiv
  preprint arXiv:2211.12445  (2022)

\bibitem{wei2022mask}
Wei, J., Zhang, Z., Shen, F., Lv, C.: Mask-guided generation method for
  industrial defect images with non-uniform structures. Machines
  \textbf{10}(12), ~1239 (2022)

\bibitem{wieler2007weakly}
Wieler, M., Hahn, T.: Weakly supervised learning for industrial optical
  inspection. In: DAGM symposium in. vol.~6 (2007)

\bibitem{xie2021segformer}
Xie, E., Wang, W., Yu, Z., Anandkumar, A., Alvarez, J.M., Luo, P.: Segformer:
  Simple and efficient design for semantic segmentation with transformers
  (2021)

\bibitem{BGAD}
Yao, X., Li, R., Zhang, J., Sun, J., Zhang, C.: Explicit boundary guided
  semi-push-pull contrastive learning for supervised anomaly detection  (2023),
  \url{https://arxiv.org/abs/2207.01463}

\bibitem{yu2021bisenet}
Yu, C., Gao, C., Wang, J., Yu, G., Shen, C., Sang, N.: Bisenet v2: Bilateral
  network with guided aggregation for real-time semantic segmentation.
  International Journal of Computer Vision  \textbf{129},  3051--3068 (2021)

\bibitem{zhang2021defect}
Zhang, G., Cui, K., Hung, T.Y., Lu, S.: Defect-gan: High-fidelity defect
  synthesis for automated defect inspection. In: Proceedings of the IEEE/CVF
  Winter Conference on Applications of Computer Vision. pp. 2524--2534 (2021)

\bibitem{AeBADindustrial}
Zhang, Z., Zhao, Z., Zhang, X., Sun, C., Chen, X.: Industrial anomaly detection
  with domain shift: A real-world dataset and masked multi-scale
  reconstruction. arXiv preprint arXiv:2304.02216  (2023)

\bibitem{zhao2017pyramid}
Zhao, H., Shi, J., Qi, X., Wang, X., Jia, J.: Pyramid scene parsing network.
  In: Proceedings of the IEEE conference on computer vision and pattern
  recognition. pp. 2881--2890 (2017)

\bibitem{zhou2017scene}
Zhou, B., Zhao, H., Puig, X., Fidler, S., Barriuso, A., Torralba, A.: Scene
  parsing through ade20k dataset. In: Proceedings of the IEEE Conference on
  Computer Vision and Pattern Recognition (2017)

\bibitem{zhou2019semantic}
Zhou, B., Zhao, H., Puig, X., Xiao, T., Fidler, S., Barriuso, A., Torralba, A.:
  Semantic understanding of scenes through the ade20k dataset. International
  Journal of Computer Vision  \textbf{127}(3),  302--321 (2019)

\bibitem{CycleGAN2017}
Zhu, J.Y., Park, T., Isola, P., Efros, A.A.: Unpaired image-to-image
  translation using cycle-consistent adversarial networks. In: Computer Vision
  (ICCV), 2017 IEEE International Conference on (2017)

\bibitem{zou2022spotthedifference}
Zou, Y., Jeong, J., Pemula, L., Zhang, D., Dabeer, O.: Spot-the-difference
  self-supervised pre-training for anomaly detection and segmentation (2022)

\end{thebibliography}
\title{Supplementary Material for Defect Spectrum: A Granular Look of Large-Scale Defect Datasets with Rich Semantics} 

\titlerunning{Defect Spectrum}

\author{Shuai Yang\inst{1,2}\thanks{These authors contributed equally to this work.} \and
Zhifei Chen\inst{1}\printfnsymbol{1}\and
Pengguang Chen\inst{3} \and
Xi Fang\inst{3} \and
Shu Liu\inst{3} \and
Yingcong Chen\inst{1,2,4}
}
\institute{Hong Kong University of Science and Technology, Guangzhou \and
HKUST(GZ) - SmartMore Joint Lab\\
 \and
SmartMore. Corp\\
\and
Hong Kong University of Science and Technology
}
\authorrunning{S.~Yang, Z.~Chen et al.}

\maketitle

In this supplementary, we extended our experiment to incorporate more annotation comparisons with existing datasets in Sec.~\ref{anno_compare}. The detailed generation settings and more quantitative analysis are discussed in Sec.~\ref{Defect_Gen}. We also include more visual cases in Sec.~\ref{visual} to demonstrate the capacity of our framework to maintain both fidelity and diversity.

\section{Visual Comparison between Original and Defect Spectrum Dataset} 
\label{anno_compare}
\begin{figure*}[htbp]
\centering
\includegraphics[width=1\linewidth]{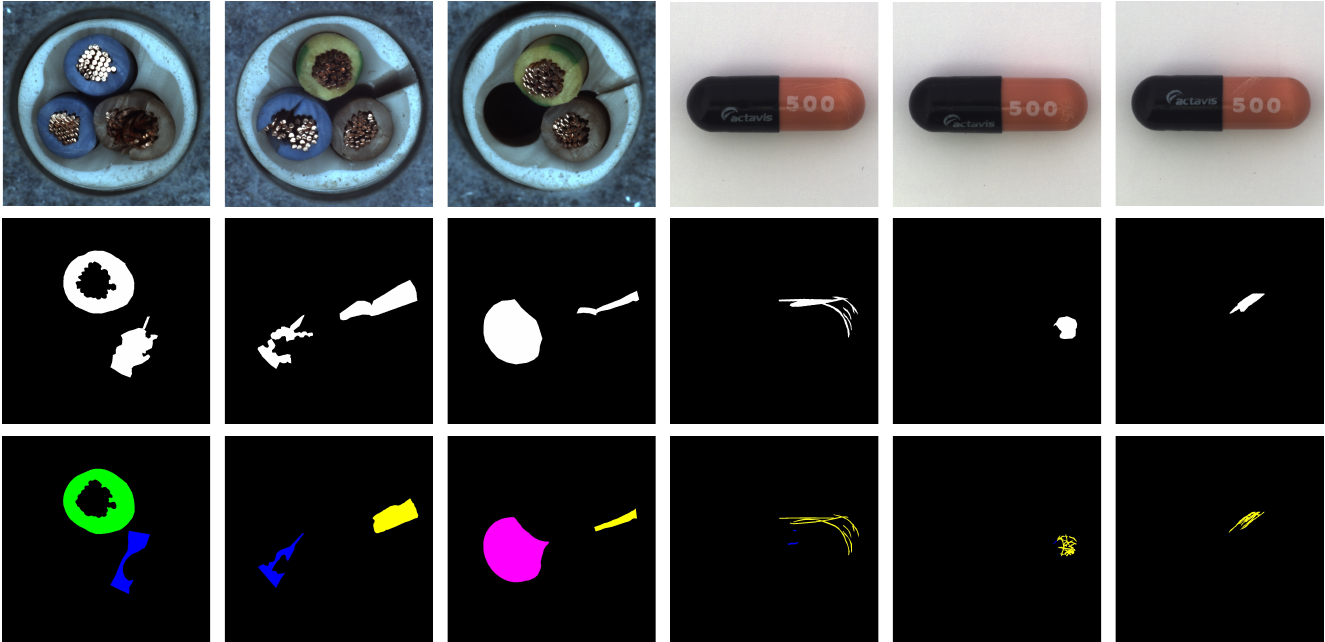}
\caption{The annotation comparison of the ``cable" and ``capsule" class in MVTec dataset. The first row shows the defect image. Rows 2 and 3 show the original annotation and our improved annotation. \textbf{Best viewed in color.}}
\label{anno_1}
\end{figure*}

\begin{figure*}[htbp]
\centering
\includegraphics[width=1\linewidth]{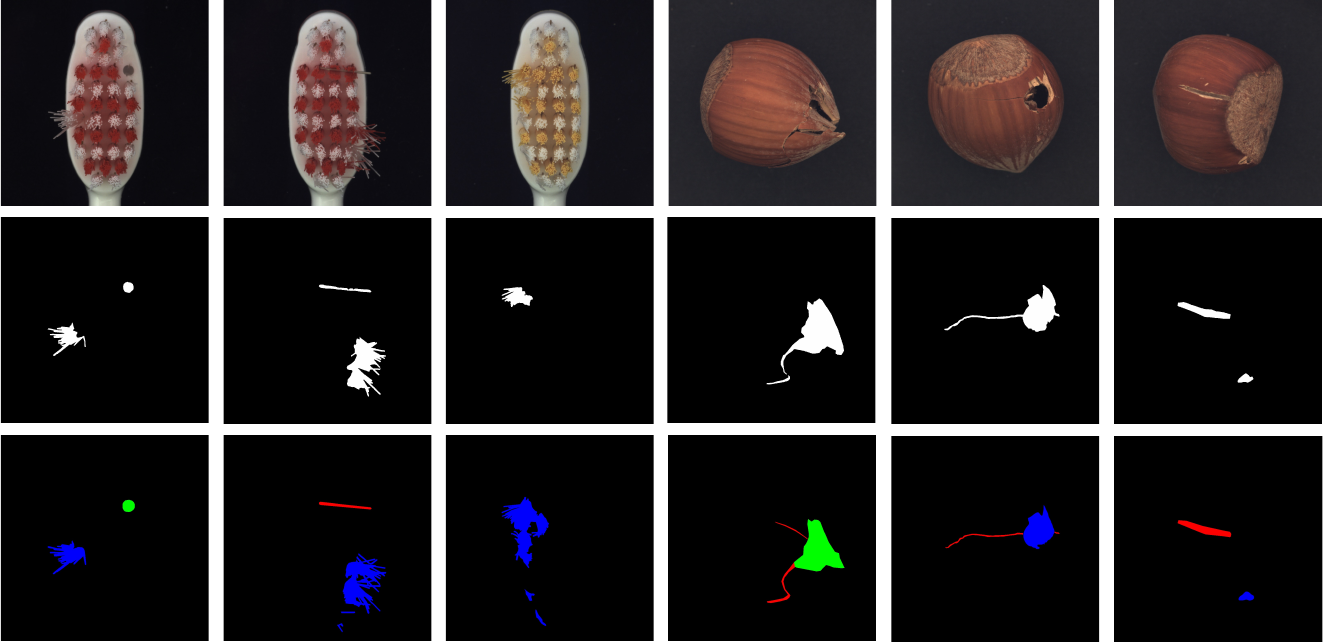}
\caption{The annotation comparison of the ``toothbrush" and ``hazelnut" class in MVTec dataset. The first row shows the defect image. Rows 2 and 3 show the original annotation and our improved annotation. \textbf{Best viewed in color.}}
\vspace{0.5in}
\label{anno_2}
\end{figure*}

\begin{figure*}[htbp]
\centering
\includegraphics[width=1\linewidth]{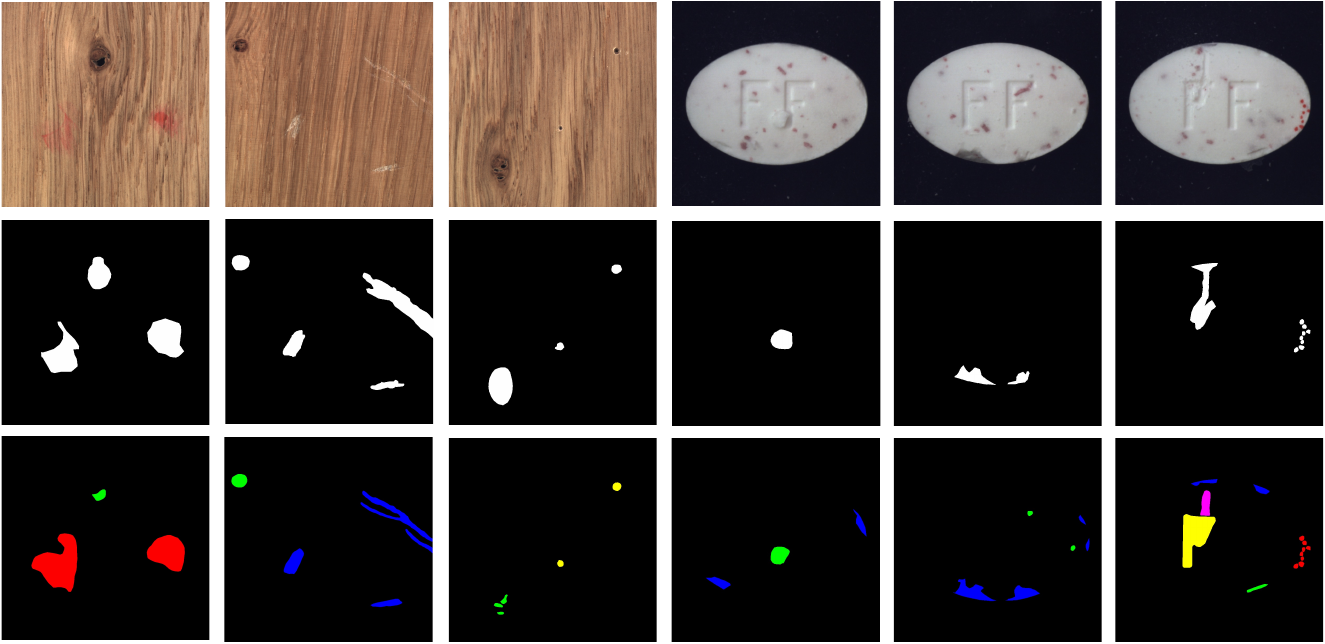}
\caption{The annotation comparison of the ``wood" and ``pill" class in MVTec dataset. The first row shows the defect image. Row 2 and 3 show the original annotation and our improved annotation. \textbf{Best viewed in color.}}
\label{anno_3}
\end{figure*}

\begin{figure*}[htbp]
\centering
\includegraphics[width=1\linewidth]{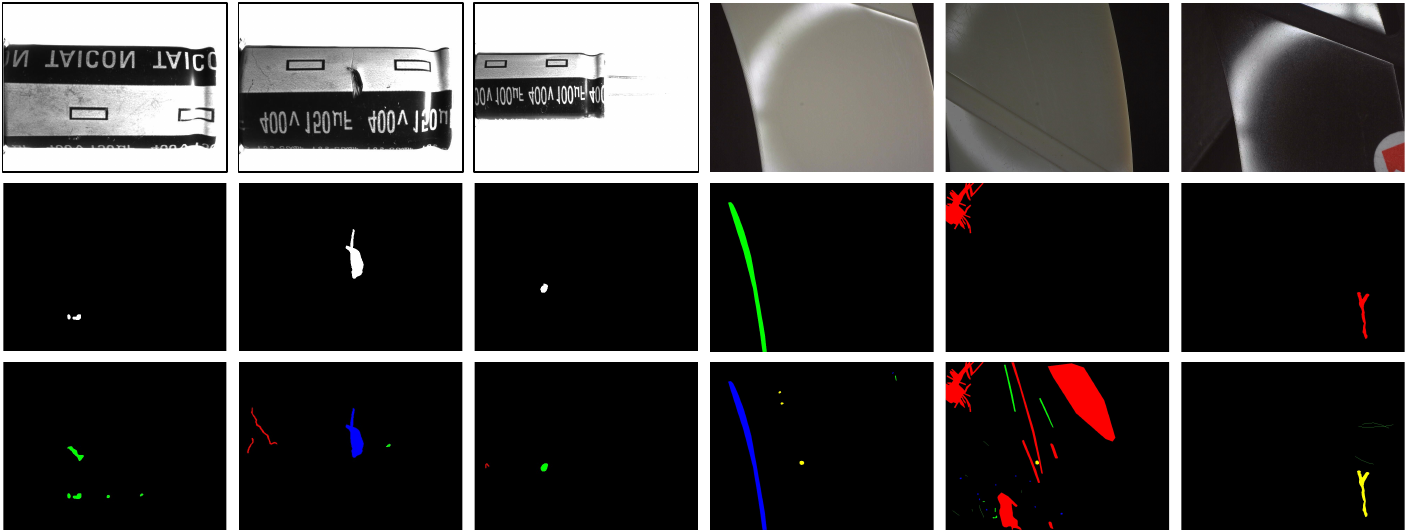}
\caption{The annotation comparison of the ``capacitor" and ``ring" class in VISION dataset. The first row shows the defect image. Rows 2 and 3 show the original annotation and our improved annotation. \textbf{Best viewed in color.}}
\label{anno_4}
\end{figure*}

\begin{figure*}[htbp]
\centering
\includegraphics[width=1\linewidth]{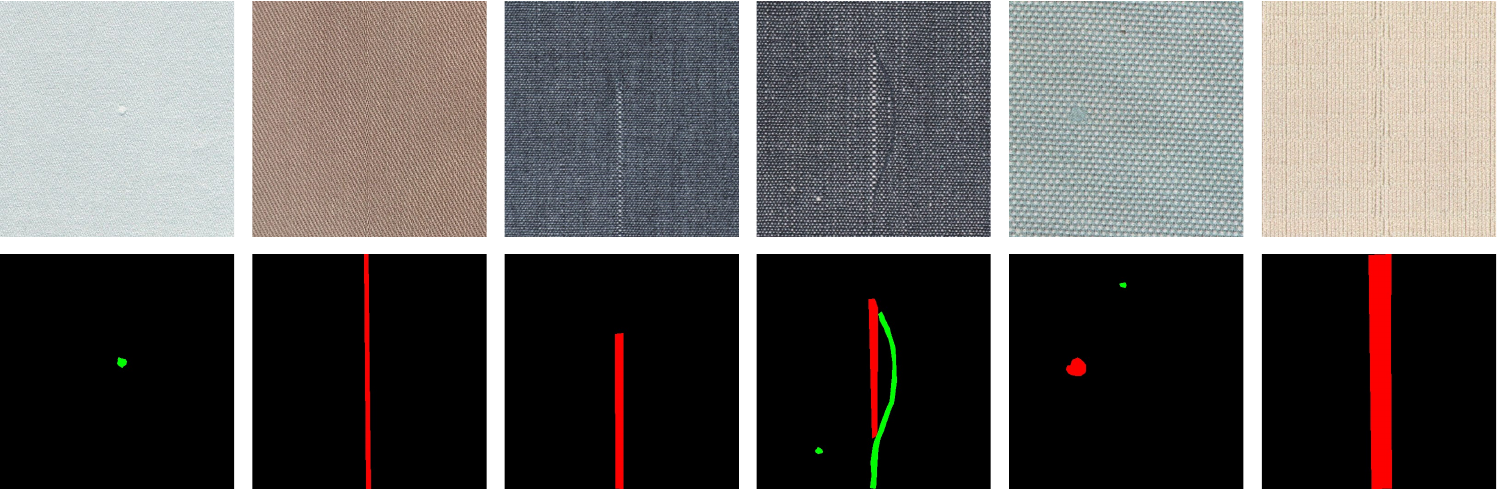}
\caption{The annotation comparison of the ``cotton fabric"  class in the COTTON dataset. The first row shows the defect image. Row 2 shows our improved annotation. \textbf{Best viewed in color.}}
\label{anno_5}
\end{figure*}

\begin{figure*}[htbp]
\centering
\includegraphics[width=1\linewidth]{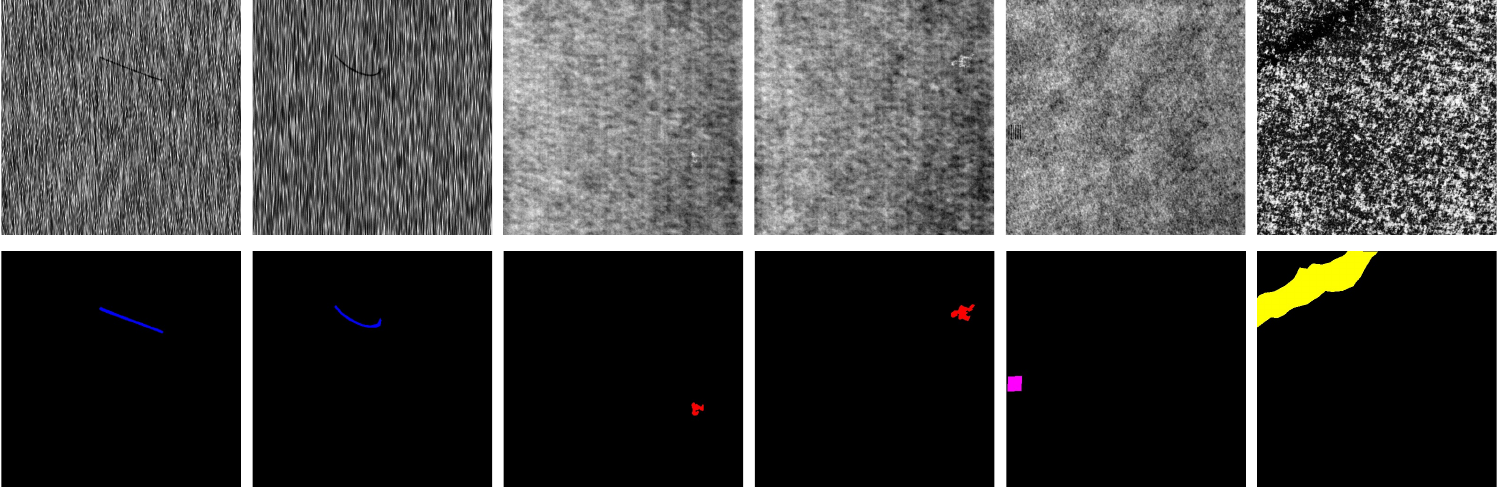}
\caption{The annotation comparison of the ``texture surface" in DAGM dataset. The first row shows the defect image. Row 2 shows our improved annotation. \textbf{Best viewed in color.}}
\label{anno_6}
\end{figure*}

\begin{figure*}[htbp]
\centering
\includegraphics[width=0.7\linewidth]{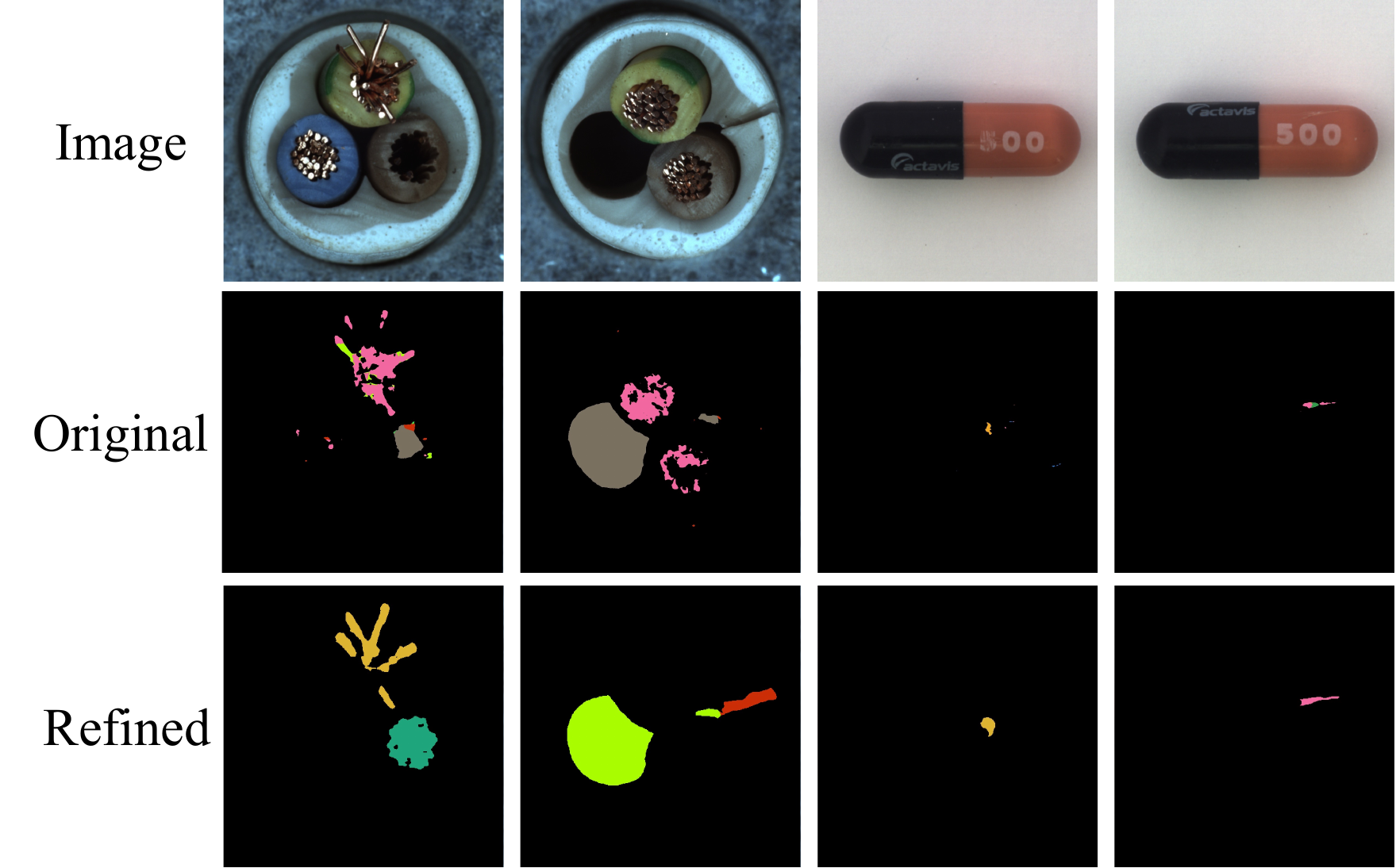}
\caption{Segmentation result comparison between model trained on our refined dataset and the original dataset of the ``cable" and ``capsule" class in MVTec dataset. ``Original'' denotes the segmentation masks produced by the model trained on the original dataset. ``Refined'' denotes the segmentation masks produced by the model trained on our refined dataset. We show the model trained with our dataset exhibits improved granularity and high quality. \textbf{Best viewed in color.}}
\label{fig:base_vs_refine_mvtec_1}
\end{figure*}

\begin{figure*}[htbp]
\centering
\includegraphics[width=0.7\linewidth]{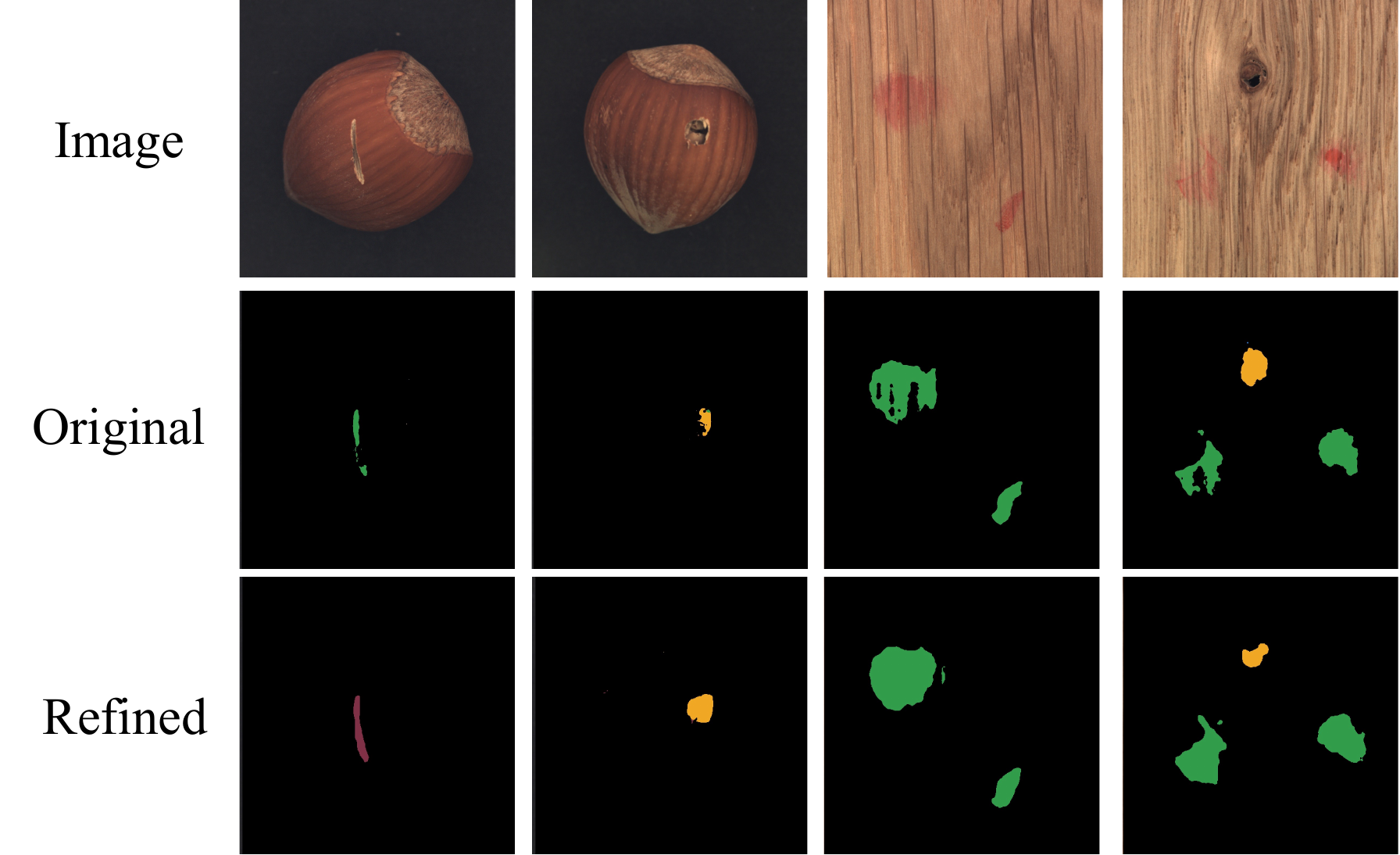}
\caption{Segmentation result comparison between model trained on our refined dataset and the original dataset of the ``hazelnut" and ``wood" class in MVTec dataset. ``Original'' denotes the segmentation masks produced by the model trained on the original dataset. ``Refined'' denotes the segmentation masks produced by the model trained on our refined dataset. We show the model trained with our dataset exhibits improved granularity and high quality. \textbf{Best viewed in color.}}
\label{fig:base_vs_refine_mvtec_2}
\end{figure*}

\begin{figure*}[htbp]
\centering
\includegraphics[width=0.7\linewidth]{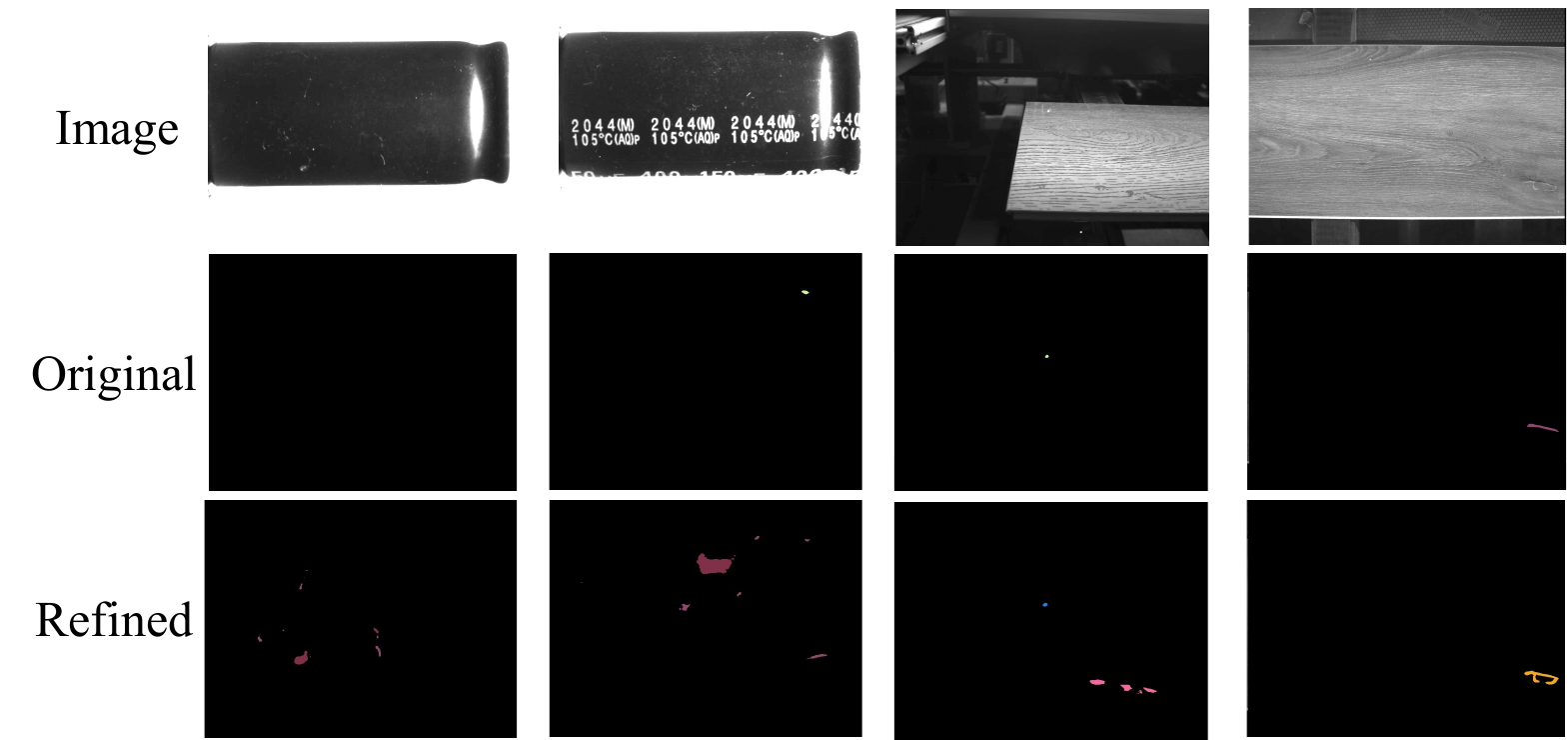}
\caption{Segmentation result comparison between model trained on our refined dataset and the original dataset of the ``Capacitor" and ``Wood" class in the VISION dataset. ``Original'' denotes the segmentation masks produced by the model trained on the original dataset. ``Refined'' denotes the segmentation masks produced by the model trained on our refined dataset. We show the model trained with our dataset exhibits improved granularity and high quality. \textbf{Best viewed in color.}}
\label{fig:base_vs_refine_vision}
\end{figure*}

In this section, we first present a visual comparison between ours (the last row) and the original datasets' annotation. Figure~\ref{anno_1}, \ref{anno_2}, \ref{anno_3} shows the comparison of the MVTec dataset, we re-classify the defects based on their type and enabled more semantic abundance. As for Figure~\ref{anno_4} of the VISION dataset, we refined the original annotation for more granularity. The original DAGM and Cotton datasets contained no pixel-level annotation, thus we provide our annotation as shown in Figure~\ref{anno_5}, \ref{anno_6}. 
We also demonstrate the efficacy of our refined annotations for defect inspection by employing a segmentation model. As illustrated in Figure~\ref{fig:base_vs_refine_mvtec_1}, \ref{fig:base_vs_refine_mvtec_2} and Figure~\ref{fig:base_vs_refine_vision}, the segmentation model trained on our refined dataset demonstrates enhanced precision and an improved capability to differentiate between various types of defects, compared to its performance when trained on the original dataset.

\vspace{1in}

\vspace{1in}

\vspace{1in}

\vspace{1in}

\section{Defect Generation}
\label{Defect_Gen}
\subsubsection{Implementation details} 
In this section, we will first elaborate on the architecture of Defect-Gen. Then we will go over the dataset and training settings of our model. Lastly, we quantitatively compared it with other methods to demonstrate the superiority of our method.

\subsubsection{Experimental Settings} Since there was no train-test split in MVTec AD dataset, to train both large and small diffusion models, we employed 5 images for each defective type per object, which is the same as our segmentation training setting. For VISION, DAGM2007, and Cotton-Fabric, we use the pre-split training set.
Table~\ref{up_block} to \ref{small_rf} show the architectures of the large and small-receptive-field models. The training of diffusion models is performed on four 3090 GPUs, with a batch size of 2, a learning rate of $1e-4$, and a training iteration number of 150,000. We utilize the Adam optimizer with a weight decay of $2e-3$.
\begin{table}[htbp]
    \caption{Upsampling Block}
     \vspace{0.1in}
  \centering
    \newcommand\widthface{0.7}
    \resizebox{1\linewidth}{!}{
    \begin{tabular}{ccccc}
    \hline
    Layer Type & Input size & Output size & Norm  & Activation \\
    \hline
    ResBlock $\times$ 2 & $H\times W\times C$ & $H\times W\times C$ & GN    & SiLU \\
    Interpolation & $H\times W \times C$ & $2H\times2W\times \frac{C}{2}$ & None  & None \\
    \hline
    \end{tabular}}

  \label{up_block}%
\end{table}%
\vspace{0.15in}
\begin{table}[htbp]
    \caption{Downsampling Block}
     \vspace{0.1in}
  \centering
      \newcommand\widthface{0.7}
    \resizebox{1\linewidth}{!}{
    \begin{tabular}{ccccc}
    \hline
    Layer Type & Input size & Output size & Norm  & Activation \\
    \hline
    ResBlock $\times$ 2 & $H\times W\times C$ & $H\times W\times C$ & GN    & SiLU \\
    Avg\_pool $2\times2$ & $H\times W\times C$ & $\frac{H}{2}\times \frac{W}{2}\times 2C$ & None  & None \\
    \hline
    \end{tabular}}

  \label{ds_block}%
\end{table}%
\vspace{0.15in}
\begin{table}[htbp]
    \caption{Architecture for Large receptive fields model.}
     \vspace{0.1in}
  \centering
      \newcommand\widthface{0.7}
    \resizebox{1\linewidth}{!}{
    \begin{tabular}{ccccc}
    \hline
    Layer Type & Resolution & \# of Channels & Norm  & Activation \\
    \hline
    InConv & 256   & 4     & GN    & SiLU \\
    DownSampleBlock & 256   & 192   & None  & None \\
    DownSampleBlock & 128   & 384   & None  & None \\
    DownSampleBlock & 64    & 768   & None  & None \\
    DownSampleBlock & 16    & 1536  & None  & None \\
    UpSampleBlock & 16    & 768   & None  & None \\
    UpSampleBlock & 64    & 384   & None  & None \\
    UpSampleBlock & 128   & 192   & None  & None \\
    UpSampleBlock & 256   & 96    & None  & None \\
    OutConv & 256   & 4     & GN    & SiLU \\
    \hline
    \end{tabular}}

  \label{large_rf}%
\end{table}%
\vspace{0.15in}
\begin{table}[htbp]
    \caption{Architecture for Small receptive fields model.}
    \vspace{0.1in}
  \centering
      \newcommand\widthface{0.7}
    \resizebox{1\linewidth}{!}{
    \begin{tabular}{ccccc}
    \hline
    Layer Type & Resolution & \# of Channels & Norm  & Activation \\
    \hline
    InConv & 256   & 4     & GN    & SiLU \\
    DownSampleBlock & 256   & 192   & None  & None \\
    DownSampleBlock & 128   & 384   & None  & None \\
    UpSampleBlock & 128   & 192   & None  & None \\
    UpSampleBlock & 256   & 96    & None  & None \\
    OutConv & 256   & 4     & GN    & SiLU \\
    \hline
    \end{tabular}}

  \label{small_rf}%
\end{table}%
\vspace{0.5in}
\subsubsection{Parameter analysis} 
As we discuss in Sec.3.4.2, our model has two key hyperparameters: the switch timestep $u$ and the receptive field of the small model. Both of them can control the trade-off between fidelity and diversity. We use FID to measure the generation fidelity. Since there is no existing metric to effectively measure the generation diversity,  we used LPIPS score to indicate such. A higher LPIPS score with a similar FID score demonstrated a higher diversity in the dataset. 
Table~\ref{tab:tradeoff} shows the FID and LPIPS for different $u$ and receptive fields. As shown, when $u$ increases, fidelity increases while diversity decreases. Similarly, when the receptive field switches from small to large, the same trend occurs. Empirically, we use $u$=50 and the medium receptive field to achieve a good trade-off between FID and LPIPS.

\begin{table}[htbp]
  \centering
    \newcommand\widthface{0.7}
    \caption{The table shows the trade-off between diversity and image quality of the capsule class. The column represents 3 different receptive field sizes, large, medium, and small, and the respective down-sampling blocks are 6, 3, 2. The row represents the timesteps($v$) used for the small receptive field model.}
    \vspace{0.15in}
    \resizebox{1\linewidth}{!}{
    \begin{tabular}{c|c|cccccc}
    \hline
    \multicolumn{1}{c}{} &  u     & 25   & 50   & 75   & 100   & 400   & 700 \\
    \hline
    Small & FID $\downarrow$   & 115.2754 & 93.2839 & 80.8040 & 79.6411 & 82.5127 & 78.4115 \\
          & LPIPS $\uparrow$& 0.3981 & 0.3666 & 0.3537 & 0.3523 & 0.3467 & 0.3460 \\
    \hline
    Medium  & FID $\downarrow$  & 69.9419 & \textcolor[rgb]{ 1,  0,  0}{57.5374} & 57.3961 & 57.8977 & 57.426 & 57.006 \\
          & LPIPS $\uparrow$& 0.3473 & \textcolor[rgb]{ 1,  0,  0}{0.3458} & 0.3450 & 0.3417 & 0.3392 & 0.3381 \\
    \hline
    Large & FID $\downarrow$  & 59.085 & 56.6246 & 56.7247 & 56.2493 & 55.7226 & 54.0529 \\
          & LPIPS $\uparrow$& 0.2914 & 0.2870 & 0.2866 & 0.2853 & 0.2832 & 0.2814 \\
    \hline
    \end{tabular}
    }  
  \label{tab:tradeoff}
\end{table}

\vspace{0.3in}
\subsubsection{Quantitative Evaluation}
We have compared the segmentation performance boost across different methods on the original MVTec dataset. GAN-based methods were excluded since they hardly generate realistic images, further disrupting the original data distribution. Results for defect segmentation are shown in Table.~\ref{tab:downstream_seg_only_baseline}.
The first column shows the defect segmentation mIoU score with only the original training data. The rest of each column presents defect segmentation performance with original training data pairs and the augmented pairs generated by different synthesis methods. SinDiffusion dropped the mIoU score, due to the incorrectly structured output images and mislabeled masks. However, it can slightly improve the segmentation performance for certain classes like ``Carpet", ``Grid", ``Leather", ``Tile" and ``Wood". Since those classes do not contain any industrial parts and thus do not require any global structure information during synthesizing.  DDPM-generated samples can boost the performance score, however, due to the lack of diversity during generation, the increase in performance is limited. 
\begin{table}[htbp]
  \centering
      \caption{Quantitative comparison on segmentation performance between sinDiffusion, DDPM, and our method. To demonstrate the effectiveness of our method on other dataset besides Defect Spectrum, the comparison was made on the original MVTec dataset }
    \begin{tabular}{c|cccc}
    \hline
    & w/o any AUG & sinDiffusion & DDPM   & Ours \\
    \hline
    capsule & 75.47 & 76.25 & 79.21 &  \textcolor[rgb]{ 1,  0,  0}{82.20} \\
    bottle & 67.54 & 70.52 & 67.32 &   \textcolor[rgb]{ 1,  0,  0}{73.75} \\
    carpet & 67.33 & 72.89 & 68.94 &  \textcolor[rgb]{ 1,  0,  0}{74.27} \\
    screw & 53.12 & 49.66 & \textcolor[rgb]{ 1,  0,  0}{60.12}  & 58.78 \\
    grid  & 59.68 & 61.59 & 60.68 &   \textcolor[rgb]{ 1,  0,  0}{62.14} \\
    cable & 46.28 & 41.75 & 48.28 &   \textcolor[rgb]{ 1,  0,  0}{49.14} \\
    hazelnut & 69.25 & 65.65 & 69.25 &   \textcolor[rgb]{ 1,  0,  0}{71.46} \\
    leather & 66.39 & \textcolor[rgb]{ 1,  0,  0}{66.91} & 66.39 &  66.80 \\
    metal\_nut & 69.56 & 63.5 & 68.57 &  \textcolor[rgb]{ 1,  0,  0}{74.4} \\
    pill  & 69.71 & 66.75 & 70.14 &  \textcolor[rgb]{ 1,  0,  0}{73.19} \\
    tile  & 70.33 & 72.43 & 71.23 & \textcolor[rgb]{ 1,  0,  0}{73.58} \\
    toothbrush & 68.26 & 64.26 & 68.09 & \textcolor[rgb]{ 1,  0,  0}{70.14} \\
    transistor & 44.31 & 47.16 & 44.37  & \textcolor[rgb]{ 1,  0,  0}{47.47} \\
    wood  & 65.33 & \textcolor[rgb]{ 1,  0,  0}{70.25} & 64.93 &  68.55 \\
    zipper & 67.62 & 63.12 & 68.61 &  \textcolor[rgb]{ 1,  0,  0}{70.48} \\

    \hline
    \textbf{mean} & 64.01  & 63.51 & 65.07  & \textcolor[rgb]{ 1,  0,  0}{67.76}\\
    \hline
    \end{tabular}
\vspace{0.5in}
  \label{tab:downstream_seg_only_baseline}%
\end{table}%
\vspace{-0.1in}




\section{Visual Generation Results} \label{visual}
We have included more defect generation results along with their masks as shown in Figure~\ref{gen_1} to \ref{gen_6} below. 
\vspace{0.3in}
\begin{figure*}[htbp]
\centering
\includegraphics[width=1\linewidth]{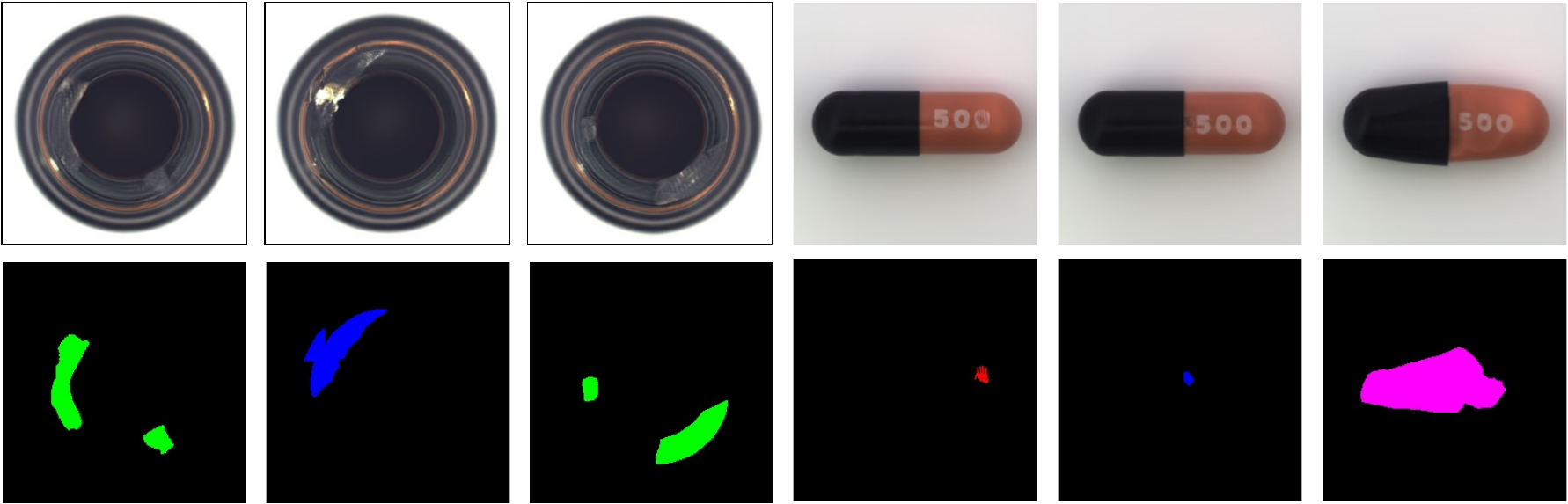}
\caption{The generated images and masks of the ``bottle" and ``capsule"  class. \textbf{Best viewed in color.}}
\vspace{0.8in}
\label{gen_1}
\end{figure*}

\begin{figure*}[htbp]
\centering
\includegraphics[width=1\linewidth]{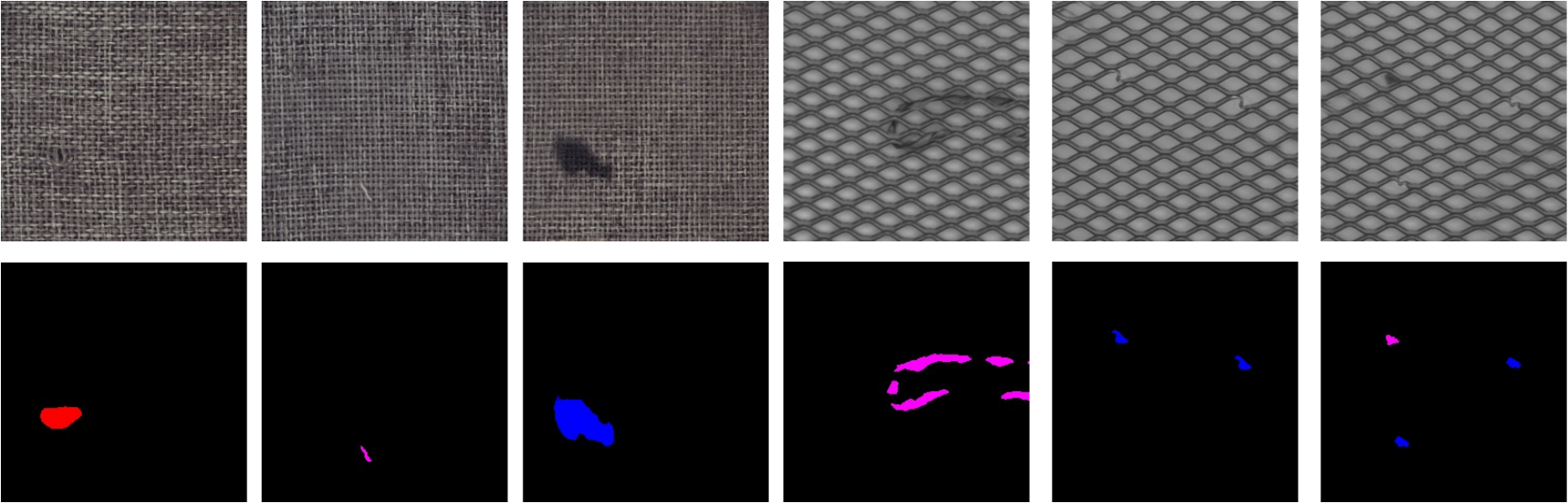}
\caption{The generated images and masks of the ``carpet" and ``grid" class. \textbf{Best viewed in color.}}
\vspace{0.5in}
\label{gen_2}
\end{figure*}

\begin{figure*}[htbp]
\centering
\includegraphics[width=1\linewidth]{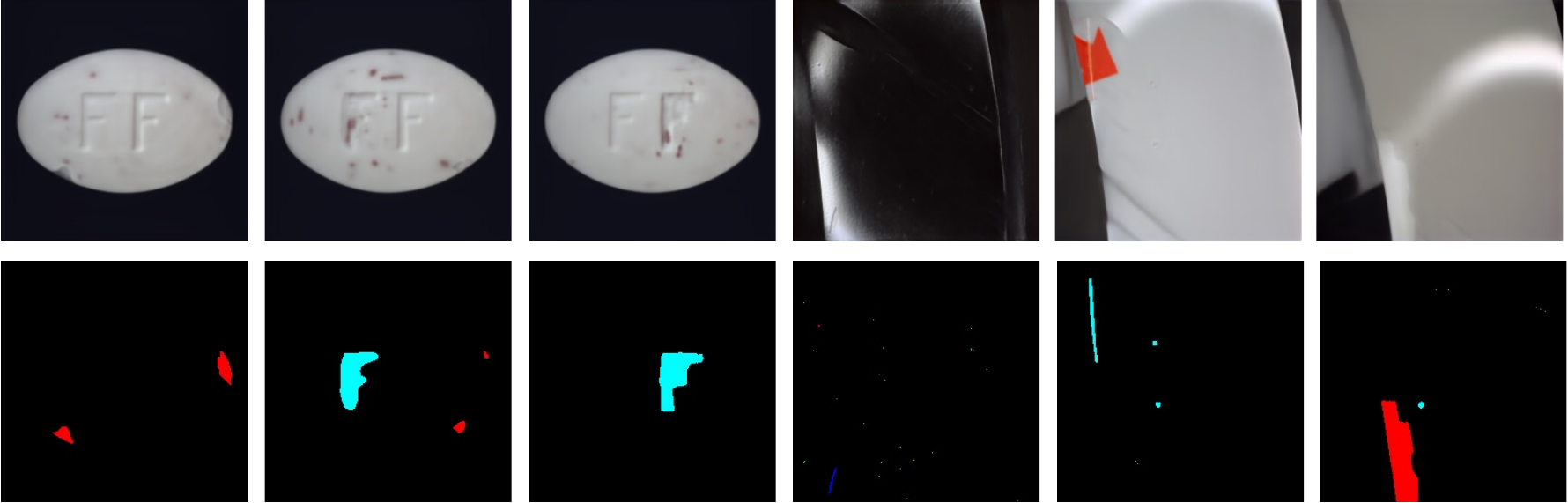}
\caption{The generated images and masks of the ``pill" and ``ring" class. \textbf{Best viewed in color.}}
\vspace{0.5in}
\label{gen_3}
\end{figure*}

\begin{figure*}[htbp]
\centering
\includegraphics[width=1\linewidth]{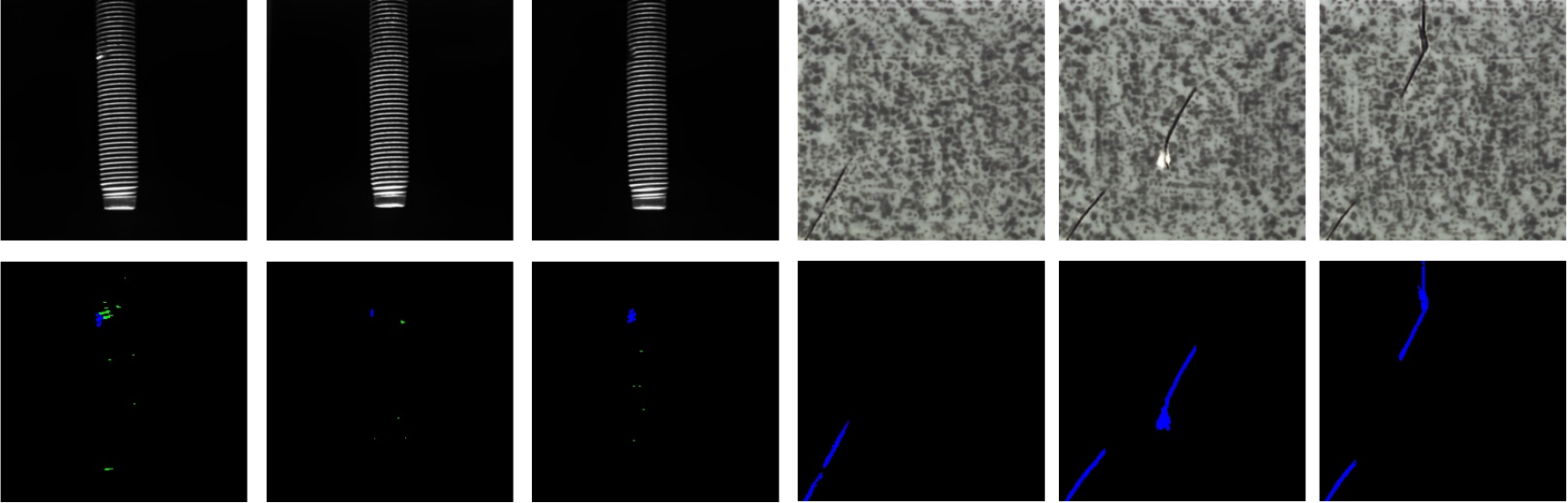}
\caption{The generated images and masks of the ``screw" and ``tile" class. \textbf{Best viewed in color.}}
\label{gen_4}
\end{figure*}

\begin{figure*}[htbp]
\centering
\includegraphics[width=1\linewidth]{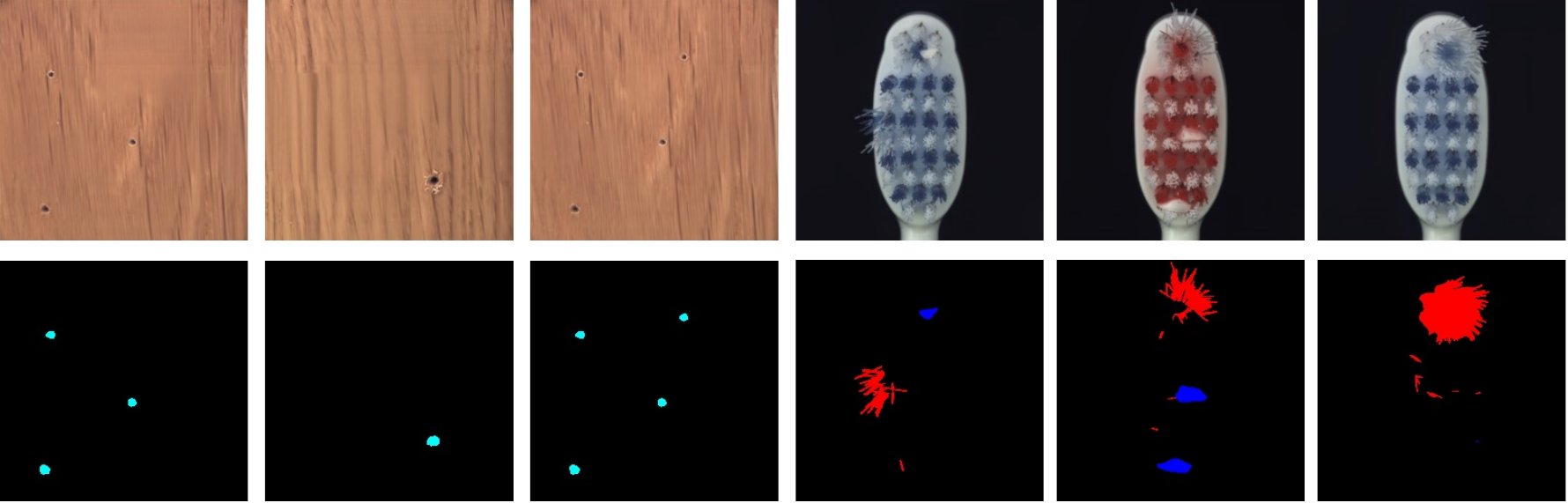}
\caption{The generated images and masks of the ``wood" and ``toothbrush" class. \textbf{Best viewed in color.}}
\label{gen_5}
\end{figure*}

\begin{figure*}[htbp]
\centering
\includegraphics[width=1\linewidth]{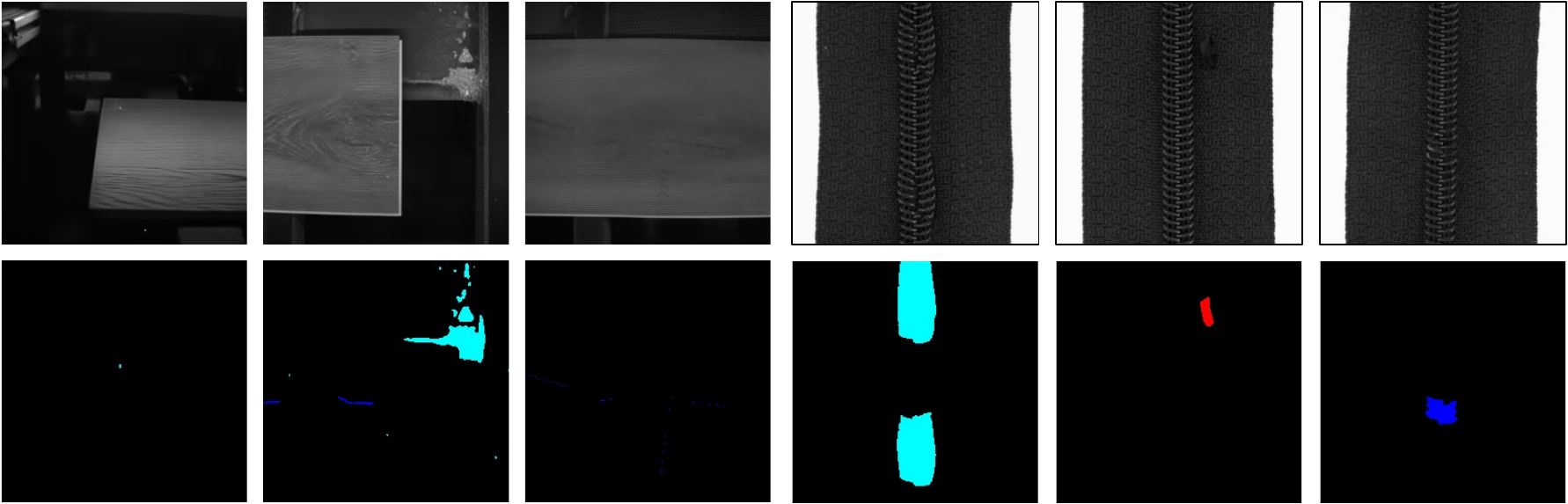}
\caption{The generated images and masks of the ``wood-surface" and ``zipper" class. \textbf{Best viewed in color.}}
\label{gen_6}
\end{figure*}


%
%
\end{document}